\documentclass{article}

\PassOptionsToPackage{sort, numbers, compress}{natbib}

\usepackage[preprint]{neurips_data_2022}

\usepackage[utf8]{inputenc} 
\usepackage[T1]{fontenc}    
\usepackage{hyperref}       
\usepackage{url}            
\usepackage{booktabs}       
\usepackage{amsfonts}       
\usepackage{nicefrac}       
\usepackage{microtype}      
\usepackage[dvipsnames]{xcolor}         
\usepackage{soul}           
\usepackage{tikz}
\usepackage{wrapfig}

\usepackage{color}
\usepackage[flushleft]{threeparttable}
\usepackage{booktabs}
\usepackage{graphicx}
\usepackage{multirow}
\usepackage{makecell}
\usepackage{xspace}
\usepackage{amssymb}
\usepackage{amsmath}
\usepackage{paralist} 
\usepackage{arydshln}
\usepackage{gensymb}
\usepackage{cinzel}
\usepackage{subcaption}
\usepackage{pdfpages}

\usepackage{listings}
\definecolor{codegreen}{rgb}{0,0.6,0}
\definecolor{codegray}{rgb}{0.5,0.5,0.5}
\definecolor{codepurple}{rgb}{0.58,0,0.82}
\definecolor{backcolour}{rgb}{0.95,0.95,0.92}
\lstdefinestyle{mystyle}{
    backgroundcolor=\color{backcolour},   
    commentstyle=\color{codegreen},
    keywordstyle=\color{magenta},
    numberstyle=\tiny\color{codegray},
    stringstyle=\color{codepurple},
    basicstyle=\ttfamily\footnotesize,
    breakatwhitespace=false,         
    breaklines=false,                 
    captionpos=b,                    
    keepspaces=true,                 
    numbers=left,                    
    numbersep=5pt,                  
    showspaces=false,                
    showstringspaces=false,
    showtabs=false,                  
    tabsize=2
}
\usepackage{pifont}

\usepackage{lipsum}
\usepackage{enumitem}

\newcommand{\dataset}{Multi-LexSum\xspace}
\newcommand{\dataseturl}{\url{https://multilexsum.github.io}\xspace}

\newcommand{\taskdtl}{$\textsc{d}\rightarrow\textsc{l}$\xspace}
\newcommand{\taskdts}{$\textsc{d}\rightarrow\textsc{s}$\xspace}
\newcommand{\taskdtt}{$\textsc{d}\rightarrow\textsc{t}$\xspace}
\newcommand{\tasklts}{$\textsc{l}\rightarrow\textsc{s}$\xspace}
\newcommand{\taskltt}{$\textsc{l}\rightarrow\textsc{t}$\xspace}
\newcommand{\taskstt}{$\textsc{s}\rightarrow\textsc{t}$\xspace}

\DeclareRobustCommand{\exsa}[1]{{\definecolor{foo}{HTML}{FFCA3A}\sethlcolor{foo}\textsuperscript{\textcolor{WildStrawberry}{[a]}}\hl{#1}}} %
\DeclareRobustCommand{\exsd}[1]{{\definecolor{foo}{HTML}{FFCA3A}\sethlcolor{foo}\textsuperscript{\textcolor{WildStrawberry}{[d]}}\hl{#1}}} %
\DeclareRobustCommand{\exse}[1]{{\definecolor{foo}{HTML}{FFCA3A}\sethlcolor{foo}\textsuperscript{\textcolor{WildStrawberry}{[e]}}\hl{#1}}} %
\DeclareRobustCommand{\exsf}[1]{{\definecolor{foo}{HTML}{FFCA3A}\sethlcolor{foo}\textsuperscript{\textcolor{WildStrawberry}{[f]}}\hl{#1}}} %
\DeclareRobustCommand{\exsg}[1]{{\definecolor{foo}{HTML}{FFCA3A}\sethlcolor{foo}\textsuperscript{\textcolor{WildStrawberry}{[g]}}\hl{#1}}} %
\DeclareRobustCommand{\exsh}[1]{{\definecolor{foo}{HTML}{FFCA3A}\sethlcolor{foo}\textsuperscript{\textcolor{WildStrawberry}{[h]}}\hl{#1}}} %
\DeclareRobustCommand{\exsi}[1]{{\definecolor{foo}{HTML}{FFCA3A}\sethlcolor{foo}\textsuperscript{\textcolor{WildStrawberry}{[i]}}\hl{#1}}} %
\DeclareRobustCommand{\exsj}[1]{{\definecolor{foo}{HTML}{FFCA3A}\sethlcolor{foo}\textsuperscript{\textcolor{WildStrawberry}{[j]}}\hl{#1}}} %
\DeclareRobustCommand{\exsk}[1]{{\definecolor{foo}{HTML}{FFCA3A}\sethlcolor{foo}\textsuperscript{\textcolor{WildStrawberry}{[k]}}\hl{#1}}} %
\DeclareRobustCommand{\exsl}[1]{{\definecolor{foo}{HTML}{FFCA3A}\sethlcolor{foo}\textsuperscript{\textcolor{WildStrawberry}{[l]}}\hl{#1}}} %
\DeclareRobustCommand{\exsm}[1]{{\definecolor{foo}{HTML}{FFCA3A}\sethlcolor{foo}\textsuperscript{\textcolor{WildStrawberry}{[m]}}\hl{#1}}} %
\DeclareRobustCommand{\exsn}[1]{{\definecolor{foo}{HTML}{FFCA3A}\sethlcolor{foo}\textsuperscript{\textcolor{WildStrawberry}{[n]}}\hl{#1}}} %
\DeclareRobustCommand{\exso}[1]{{\definecolor{foo}{HTML}{FFCA3A}\sethlcolor{foo}\textsuperscript{\textcolor{WildStrawberry}{[o]}}\hl{#1}}} %
\DeclareRobustCommand{\exsp}[1]{{\definecolor{foo}{HTML}{FFCA3A}\sethlcolor{foo}\textsuperscript{\textcolor{WildStrawberry}{[p]}}\hl{#1}}} %

\title{\dataset: Real-World Summaries of Civil Rights Lawsuits at Multiple Granularities}

\author{Zejiang Shen$^{\dagger}$ \quad 
        Kyle Lo$^{\dagger}$\quad 
        Lauren Yu$^{\diamondsuit}$ \quad 
        Nathan Dahlberg \\
        {\bf
        Margo Schlanger$^\diamondsuit$ \qquad
        Doug Downey$^{\dagger,\clubsuit}$ \quad 
        }
        \\ [1mm]
        $^\dagger$Allen Institute for AI \quad $^\diamondsuit$University of Michigan \quad $^\clubsuit$Northwestern University
        \\
        {\tt\small \{shannons, kylel, dougd\}@allenai.org}\\
        {\tt\small \{laurenyu, mschlan\}@umich.edu nadahlberg@gmail.com}
}

\begin{document}

\maketitle

\begin{abstract}

With the advent of large language models, methods for abstractive summarization have made great strides, creating potential for use in applications to aid knowledge workers processing unwieldy document collections. One such setting is the Civil Rights Litigation Clearinghouse (CRLC),\footnote{\url{https://clearinghouse.net}.} which posts information about large-scale civil rights lawsuits, serving lawyers, scholars, and the general public. Today, summarization in the CRLC requires extensive training of lawyers and law students who spend hours per case understanding multiple relevant documents in order to produce high-quality summaries of key events and outcomes. Motivated by this ongoing real-world summarization effort, we introduce \dataset, a collection of 9,280 expert-authored summaries drawn from ongoing CRLC writing. \dataset presents a challenging multi-document summarization task given the length of the source documents, often exceeding two hundred pages per case. Furthermore, \dataset is distinct from other datasets in its multiple target summaries, each at a different granularity (ranging from one-sentence ``extreme'' summaries to multi-paragraph narrations of over five hundred words). We present extensive analysis demonstrating that despite the high-quality summaries in the training data (adhering to strict content and style guidelines), state-of-the-art summarization models perform poorly on this task. We release \dataset for further research in summarization methods as well as to facilitate development of applications to assist in the CRLC's mission.\footnote{The dataset can be accessed at \dataseturl.}

\end{abstract}

\section{Introduction}

Automatic summarization is a longstanding goal of natural language processing. Recently, abstractive summarization methods powered by large pretrained language models have shown impressive results \cite{lewis2019bart,zhang2020pegasus}---raising the question of whether these methods can help real-world summarization workloads currently performed by human experts.
In this paper, we present a new dataset, \dataset, for studying automatic summarization in an important real-world application setting found in the Civil Rights Litigation Clearinghouse (CRLC).  The CRLC currently collects and presents documents and information from modern large-scale civil rights lawsuits in a manner easily understood by legal practitioners and scholars and the general public alike~\cite{clearinghouseaboutpage}. Today, the Clearinghouse relies on human legal experts to write summaries of civil rights cases, explaining their events and outcomes. This cognitively demanding task requires summary writers to comprehend multiple documents of different types (often totaling over two hundred pages of text per case); extract entities, events, and their interrelationships; and synthesize this information into a summary that captures the key details in each case's timeline. For a typical summary, this process takes an expert 1-4 hours. And it needs to be repeated as the case proceeds through the legal system, to keep the summary up-to-date.

Success in summarization automation would allow the Clearinghouse and other efforts like it to greatly increase their coverage and update their summaries in close to real time.
Quicker and less costly narrative description of important and routine lawsuits would benefit both the legal field and the general public by increasing access to and understanding of disputes and their resolutions.

We release \dataset, an abstractive summarization dataset for federal U.S. large-scale civil rights lawsuits drawn from the CRLC.
It consists of about 40,000 source documents and 9,000 expert-written summaries (covering about half as many cases). 
Besides its potential to enable new summarization capabilities to benefit the CRLC effort and others like it, \dataset has unique characteristics that make it an interesting object of study for summarization research more broadly:


\begin{itemize}[nosep, leftmargin=1em]
    \item Unlike some summarization workloads, the CRLC task requires production of summaries at {\bf multiple target levels of granularity}: tiny (25 words, on average), short (130 words), and long (650 words). Variable granularity can be valuable in many applications---e.g., short summaries are ideal to scan for items of interest, and longer summaries to explore more deeply. But to our knowledge, \dataset is the first dataset to provide summaries at multiple levels of granularity. It enables study of multi-task methods that learn from supervision at multiple granularities, and that provide controllable generation at a specified granularity, as our experiments explore.
    \vspace{0.05in}
    \item Other multi-document summarization datasets offer only 800-8,000 words in the source documents, on average \cite{deyoung_ms2_2021,fabbri_multi-news_2019,lu2020multi}, even though many applications require summarizing {\bf large collections of multiple documents}. In \dataset, the average source length is over 75,000 words.
    \vspace{0.05in}
    \item Unlike other summarization datasets that are (semi-)automatically curated \cite{hermann_teaching_2015,narayan_dont_2018,cohan_discourse-aware_2018,kryscinski_booksum_2021,grusky_newsroom_2018,sharma_bigpatent_2019,huang_efficient_2021}, \dataset consists of {\bf expert-authored summaries}.  The experts---lawyers and law students---are trained to follow carefully created guidelines, and their work is reviewed by an additional expert to ensure quality (see Appendix~\ref{sec:guidelines}).  This provides high-quality supervision and evaluation and reduces the risk of training on summaries containing facts unsupported by the source text, which can contribute to model hallucination~\cite{king2022don,maynez2020faithfulness}.
\end{itemize}

We conduct a series of experiments on \dataset, and find that existing summarization models perform poorly. 
Human assessments of model output result in an average rating of 0.43 on a 0-3 scale, showing that significant improvements are needed before the summaries can provide utility for the CRLC project. Finally, multi-task approaches that train on the multiple granularities of summaries in \dataset demonstrate promise for improving long summary quality.



\begin{table}[t]
    \small
    \resizebox{1.\linewidth}{!}{
        \begin{threeparttable}
            \renewcommand{\arraystretch}{1.20}
            \caption{Three different summaries for one case in \dataset. We highlight and label spans of text according to which fact it covers.
            }
            \label{table:example-summary}
            \begin{tabular}{llll}
            \toprule
            \multicolumn{4}{p{1.\linewidth}}{
                {\textbf{Long Summary} $\textsc{l}$: This case is about an apprenticeship test that had a disparate impact on Black apprenticeship applicants. 
                The \exsa{Equal Employment Opportunity Commission (EEOC)} filed this lawsuit on \exsd{December 27, 2004}, in \exse{U.S. District Court for the Southern District of Ohio}. 
                Filing \exsa{on behalf of thirteen Black individuals} and \exsf{similarly situated Black apprenticeship test takers}, the EEOC alleged that \exsg{the individuals' employer, the Ford Motor Company, as well as their union, the United Automobile, Aerospace, and Agricultural implement workers of America (the "UAW"), and the Ford-UAW Joint Apprenticeship Committee}, violated \exsh{Title VII of the Civil Rights Act, 42 U.S.C. \S 1981, and Michigan state anti-discrimination law}.
                At issue were the selection tests for apprenticeship training programs, whose disparate impact denied Black applicants eligibility and admission. 
                \exsi{The EEOC sought injunctive relief, as well as damages (including backpay) for the Black apprenticeship applicants.} 
                The case was assigned to \exsj{Judge Susan J. Dlott}.
                \vspace{1mm}\newline
                \exsk{The individuals also brought a separate class action against Ford and the UAW, Robinson v. Ford Motor Company, (No. 1:04-cv-00844), and the cases were consolidated on January 6, 2005}. 
                As a result, the case was transferred to \exsj{Judge S. Arthur Spiegel}.
                Six months later, in June 2005, both cases were resolved via a \exsl{class} settlement agreement. 
                Ford agreed to pay \$8.55 million and to implement a new selection process for its apprenticeship programs. 
                This agreement further required Ford to hire an industrial psychologist to design this new selection process and to place 279 members of the settlement class on the eligibility list for the Ford apprenticeship program.
                \vspace{1mm}\newline
                On \exsm{June 15, 2005}, the court found that the proposed settlement agreement was fair. 
                \exsn{2005 WL 5253339}. 
                The next day, the court ordered that Ford pay \$1.1 million to cover attorneys' fees and expenses incurred during settlement negotiations, and \$567,000 to cover fees and expenses associated with the implementation and monitoring of the settlement agreement. 
                \exsn{2005 U.S. Dist. LEXIS 12071}. 
                As \exso{the settlement was initially scheduled to last for three years}, and \exsp{there is no further activity on the docket sheet}, this case presumably closed in 2008.}} \\
            \midrule
            \multicolumn{4}{p{1.\linewidth}}{
                {\small \textbf{Short Summary} $\textsc{s}$: This case is about an apprenticeship test that had a disparate impact on Black apprenticeship applicants. The Equal Employment Opportunity Commission (EEOC) filed this lawsuit on December 27, 2004, in U.S. District Court for the Southern District of Ohio. Filing on behalf of thirteen Black individuals and similarly situated Black apprenticeship test takers, the EEOC alleged that the individuals' employer, the Ford Motor Company, as well as their union, the United Automobile, Aerospace, and Agricultural implement workers of America (the ``UAW''), and the Ford-UAW Joint Apprenticeship Committee, violated Title VII of the Civil Rights Act, 42 U.S.C. § 1981, and Michigan state anti-discrimination law. The EEOC sought injunctive relief and damages for the Black apprenticeship applicants. The individuals also brought a separate class action against Ford and the UAW, and the cases were consolidated. In June 2005, both cases were resolved via a class settlement agreement. Ford agreed to pay \$8.55 million and to implement a new selection process for its apprenticeship programs, and the court ordered Ford to cover attorneys' fees and expenses. This case is closed.}}\\
            \midrule
            \multicolumn{4}{p{1.\linewidth}}{
                {\small \textbf{Tiny Summary} $\textsc{t}$: 2005 class action settlement resulted in Ford paying \$8.55m to redesign its selection process for apprenticeship programs to address the previous process's disparate impact on Black applicants.}}\\
            \midrule
            \multicolumn{4}{c}{\textbf{Checklist of Facts during Writing}} \\
            a. Plaintiff description     & 
            e. Court's full name         & 
            i. Remedy sought             & 
            m. Date of settlement/decree \\
            b. Type of counsel           & 
            f. Class description         & 
            j. Judge's name              & 
            n. Citation to an opinion    \\
            c. Type of action            & 
            g. Defendant description     & 
            k. Consolidated case         & 
            o. How long decrees lasted   \\
            d. Filling Date              & 
            h. Statutory basis for case  & 
            l. If class action           & 
            p. Last action in case       \\
            \bottomrule
            \end{tabular}
    \end{threeparttable}
    }
    \vspace{-5mm}
\end{table}

\section{Related work}

\subsection{Natural language processing for legal documents}

Much recent work in Natural Language Processing (NLP) has focused on the legal domain~\cite{bommarito_ii_lexnlp_2018, chalkidis_lexglue_2022, bommasani_opportunities_2021, xiao_lawformer_2021, chalkidis_legal-bert_2020}.
Lawsuits generate rich document sets with domain-specific language and complex structures, which are challenging for state-of-the-art language processing models~\cite{bommasani_opportunities_2021}.  Given the important societal role of litigation, along with the extremely high cost of legal expertise, NLP methods to help search, synthesize, and answer questions about legal corpora are of strong interest.

NLP has been applied to a variety of legal document types, including patents~\cite{sharma_bigpatent_2019}, legal provisions and contracts~\cite{tuggener-etal-2020-ledgar, lippi2019claudette, ravichander-etal-2019-question}, legislative bills~\cite{kornilova_billsum_2019}, and court documents~\cite{nallapati_legal_nodate,zheng_when_2021,grover-etal-2004-holj}.
The NLP tasks studied in this work range from document/sentence classification~\cite{chalkidis-etal-2019-neural,tuggener-etal-2020-ledgar, chalkidis2021multieurlex} to information extraction~\cite{chalkidis-etal-2018-obligation, hendrycks2021cuad}, question answering~\cite{kim2015convolutional, ravichander-etal-2019-question, kien-etal-2020-answering, zhong2020jec}, and---most relevant to our work---automatic summarization~\cite{kornilova_billsum_2019, sharma_bigpatent_2019, grover-etal-2004-holj, jain2021summarization}.
As found in other specialized domains of language, legal NLP systems often benefit from starting from a large language model pre-trained on legal text~\cite{xiao_lawformer_2021, chalkidis_legal-bert_2020,zheng_when_2021}.  

Our \dataset dataset is focused on automatic summarization of court proceedings and outcomes. Previous work on this task mainly focuses on {\em extractive} approaches, where the output summaries consist of sentences drawn directly from the source~\cite{jain2021summarization}. 
\citet{hachey2006extractive} summarize UK court judgments from the HOLJ corpus~\cite{grover-etal-2004-holj} by selecting the most summary-worthy sentences from a document, while \citet{kim2012summarization} develop a graph-based summary sentence selection method on the same corpus. 
\citet{yousfi2010supervised} propose ProdSum, a Naive Bayes sentence classifier, for summarizing case decisions from the Canadian Legal Information Institute (CanLII). 
\citet{galgani2014hauss} experiment with citation based summarization approaches on case reports from the Australasian Legal Information Institute AustLII~\cite{greenleaf1995public}. 
Systems like CaseSummarizer~\cite{polsley_casesummarizer_nodate}, LetSum~\cite{farzindar2004letsum}, and the pipeline by~\citet{zhong2019automatic} are developed to extractively summarize documents from AustLII, CanLII, and Board of Veterans' Appeals (BVA), respectively. 
By contrast, our work focuses on \emph{abstractive} summarization, where the target output is a paraphrase of the source, creating the potential for providing more succinct summaries in more accessible language. We release a large dataset of over 9,000 expert-written summaries for court documents from about 4,500 U.S. federal civil rights lawsuits, as a testbed for state-of-the-art abstractive summarizers.


\subsection{Summarization datasets in other domains}

\dataset contains expert-written summaries of up to three different granularities for the same source; this is the first such published dataset to the best of our knowledge. 
Perhaps the most similar work is BookSum~\cite{kryscinski_booksum_2021}; however in contrast to our work, BookSum's multiple summaries consider different lengths of the source to be summarized---paragraphs, chapters, and the whole content in a book.  \dataset presents a new opportunity to study how to learn from and produce summaries at varying granularity for the same source, as we explore in our experiments.

Another key differentiating factor in \dataset is that its summaries are expert-provided. 
In order to scale to impressive sizes, many existing summarization datasets are created in a (semi-)automatic fashion---e.g., using the first sentence~\cite{narayan_dont_2018} or summary bullets~\cite{nallapati_abstractive_2016} as the target summary for a piece of news, or automatically extracting and linking scientific paper abstracts~\cite{cohan_discourse-aware_2018} and citing sentences~\cite{deyoung_ms2_2021}.  
These datasets lack a clear specification of how the summary corresponds to the source, can have varying quality, and often contain information that is not directly supported or implied by the source, which can degrade the factual consistency of models trained on the data~\cite{king2022don,maynez2020faithfulness}. 
By contrast, \dataset contains ``gold'' summaries.
Experts are specifically trained to write the case summaries following carefully crafted instructions (detailed in Appendix~\ref{sec:guidelines}), and the written summaries are subsequently reviewed to ensure correctness and stylistic consistency.

Compared to many existing single- or multi-doc summarization datasets for news~\cite{narayan_dont_2018, nallapati_abstractive_2016, fabbri2019multi}, scientific papers~\cite{cohan_discourse-aware_2018,deyoung_ms2_2021,lu2020multi}, patents~\cite{sharma_bigpatent_2019}, legislative bills~\cite{kornilova_billsum_2019}, and government reports~\cite{huang_efficient_2021}, the summary context in \dataset comes from multiple sources that are  extraordinarily long---over 75k words, an order of magnitude larger than most other datasets (see Table~\ref{table:overall-datasets}). 
One exception is BookSum~\cite{kryscinski_booksum_2021}, which uses entire books as summary inputs; the books are on average 127,000 words long.  However, it has far fewer samples (403) than \dataset does (4,500).

\section{\dataset}
\label{sec:dataset}

\subsection{Task definition}

In the American legal system, civil lawsuits (``cases'') involve a set of actions among two or more parties and the judge(s)~\cite{uscourtscivilcases}. Most steps in the case are taken by way of formal document filings.
The first step typically occurs when the ``plaintiffs''---people, groups of people, or entities---file a ``complaint'' against one or more ``defendants'' in a state or federal trial court. The case then proceeds as the parties file additional documents.
It is through these documents that the parties lay out the case background, explain their arguments, rebut opposing parties' arguments, and ask for specific actions from the judge(s) (see Table~\ref{table:document-types} for a breakdown of document types).
The judge(s) also file documents which set schedules, ask questions, and memorialize rulings---intermediate orders that frame the conflict or instruct parties to take various steps or "final" orders that at least temporarily resolve the case~\cite{uscourtscoveringcivilcases}. 
All a result, a case's documents can extend to hundreds, even thousands, of pages of text.
Collectively, these documents paint a full picture of the case, but they can be extremely time-consuming to read and digest in order to understand the gist.

The goal of legal case summarization is to write a short article that captures principal details and describes each case's litigation history in plain language---information that is otherwise often difficult to come by.
The CRLC summaries come in three different lengths: \begin{itemize}[nosep,leftmargin=1em,labelwidth=*,align=left]
  \item \textbf{Long} ($\textsc{l}$) summaries typically contain multiple paragraphs, covering the case background, parties involved, and proceedings. Major case events and outcomes typically receive a paragraph each. 
  \item \textbf{Short} ($\textsc{s}$) summaries have only one paragraph with a shorter description of the background, parties involved, and the outcome (so far) of the case. 
  \item \textbf{Tiny} ($\textsc{t}$) summaries are one-sentence summaries intended to appear on Twitter to describe the case at a specific point in its history.
\end{itemize}

\begin{figure}[t]
    \centering
    \includegraphics[width=0.85\linewidth]{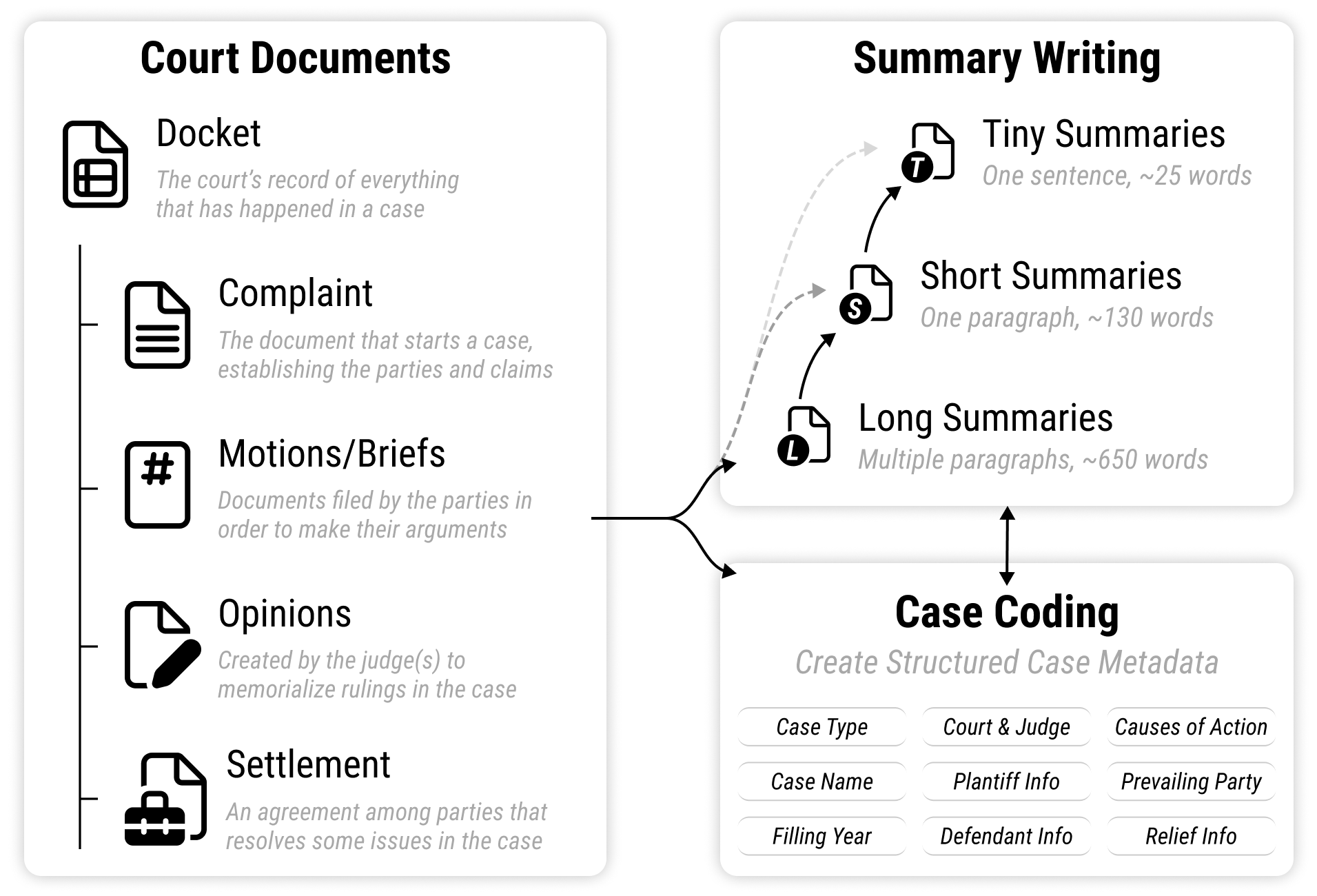}
    \caption{The pipeline of creating summaries and structured data for lawsuits.} 
    \label{fig:summary-writing}
    \vspace{-3mm}
\end{figure}
\begin{table}[t]
\caption{Comparison of \dataset to other single-document (SDS) and multi-document (MDS) summarization datasets. Measurements include dataset size, number of source documents per sample, number of words and sentences in source and target texts, and source-target coverage, density, and compression ratio. Except for number of samples, all reported values are averages across all samples, including test sets when available.
} 
\label{table:overall-datasets}
  \resizebox{1.\linewidth}{!}{
  \begin{threeparttable}
    \renewcommand{\arraystretch}{1.15}
    \setlength{\tabcolsep}{4pt}
    \begin{tabular}{rrrrrrrrrrrr}
    \toprule
       &
       &
       \multicolumn{3}{c}{\textbf{Source}} &
       &
       \multicolumn{2}{c}{\textbf{Target}} &
       &
       \multicolumn{3}{c}{\textbf{Source \textrightarrow  Target}} \\
       \cmidrule{3-5} \cmidrule{7-8} \cmidrule{10-12}   
     \textbf{Dataset} &
       \textbf{Samples} &
       Docs &
       Words &
       Sents &
       &
       Words &
       Sents &
       &
       Coverage &
       Density &
       Compress \\
\midrule
\multicolumn{12}{c}{\textbf{Tiny}}                                                               \\
\midrule
\textbf{XSUM}                      & 226,677   & 1  & 454.5    & 31.7   &  & 24.0   & 1.0   &  & 0.67 & 1.10  & 19.9   \\
\textbf{SciTLDR}                   & 3,229     & 1  & 5847.7   & 232.3  &  & 22.2   & 1.1   &  & 0.95 & 4.85  & 310.8  \\
\textbf{Newsroom}                  & 1,212,739 & 1  & 800.1    & 37.9   &  & 31.2   & 1.5   &  & 0.83 & 9.53  & 43.6   \\
\textbf{BookSum/Paragraph}\tnote{1} & 147,665 & 1 & 163.131 & 8.411 &  & 35.8 & 1.9 &  & 0.51 & 0.90 & 6.7 \\
\cdashline{1-12}[.4pt/1pt]\noalign{\vskip 1pt}
\textbf{\dataset} \taskdtt           & 1,603     & 10.7  &   119072.6 &  5962.5 &  & 24.7   & 1.4   &  & 0.92 & 2.27  & 5449.6 \\
\midrule
\multicolumn{12}{c}{\textbf{Short}}                                                                          \\
\midrule
\textbf{BigPatent}                 & 1,341,362 & 1  & 3629.0   & 131.4  &  & 116.7  & 3.5  &  & 0.86 & 2.38  & 36.8   \\
\textbf{MS\string^2}               & 16,212    & 24.0 & 7775.6   & 306.3  &  & 65.1   & 3.9  &  & 0.86 & 1.91  & 174.8  \\
\textbf{Multi-XScience}            & 40,528    & 5.1  & 817.0    & 32.0   &  & 119.7  & 4.9  &  & 0.67 & 1.30  & 7.7    \\
\textbf{CNN / Daily Mail}          & 311,971   & 1  & 805.2    & 39.3   &  & 59.9   & 6.1  &   & 0.85 & 3.49  & 14.9   \\
\textbf{BillSum}                   & 23,455    & 1  & 1804.1   & 54.4   &  & 218.4  & 6.4   &  & 0.90 & 4.05  & 12.9   \\
\cdashline{1-12}[.4pt/1pt]\noalign{\vskip 1pt}
\textbf{\dataset} \taskdts           & 3,138     & 10.3  &   99378.2     &  5017.0      &  & 130.2  & 5.1  &  & 0.96 & 3.33  & 840.7  \\
\midrule
\multicolumn{12}{c}{\textbf{Long}}                                                                        \\
\midrule
\textbf{Multi-News}                & 56,216    & 2.8  & 2168.1   & 92.2   &  & 264.0  & 10.4 &  & 0.83 & 5.01  & 8.2    \\
\textbf{BookSum/Chapter} & 12,570 & 1 & 5339.6 & 302.1 &  & 421.0 & 21.7 &  & 0.78 & 1.47 & 16.6 \\
\textbf{BookSum/Book} & 403 & 1 & 126537.2 & 6964.2 &  & 1163.1 & 56.0 &  & 0.90 & 1.79 & 146.3 \\
\cdashline{1-12}[.4pt/1pt]\noalign{\vskip 1pt}
\textbf{\dataset} \taskdtl           & 4,534     & 8.8  & 75543.2  & 3814.2 &  & 646.5  & 28.8 &  & 0.94 & 4.07  & 97.4   \\
\bottomrule
      \end{tabular}
      \begin{tablenotes}
        \item[1] The BookSum number might be slightly different from those reported in original paper because some samples weren't successfully downloaded using the script provided by the authors.
    \end{tablenotes}
  \end{threeparttable}
  }
  \vspace{-3mm}
\end{table}

Using the different granularities of summary, we define a variety of distinct summarization tasks.  First, we consider three different multi-document summarization (MDS) tasks that map from the source documents $\textsc{d}$ to each of the summary lengths above (e.g., $\textsc{d}\rightarrow\textsc{l}$ denotes the task of mapping the source documents to the long summary).  We also consider three different single-document summarization (SDS) tasks that take a ground truth summary as input and attempt to map to a shorter summary as output (e.g., $\textsc{s}\rightarrow\textsc{t}$ denotes mapping from a short summary to the corresponding tiny one). Finally, the multiple granularities in \dataset create the opportunity to use {\em sets} of the data as input or output, which we also explore (e.g. $\{\textsc{l}, \textsc{d}\}\rightarrow\textsc{t}$ denotes taking a long summary and the source documents as input, and outputting a tiny summary).

\subsection{Creating \dataset summaries}

All the data in \dataset, including the selected documents, summaries, and the structured case metadata, are manually curated by legal experts: 
legal scholars, attorneys, and
law students who 
receive specialized training relevant to their CRLC assignments.
For a case where not much has happened since the lawsuit was filed, it typically takes one to two hours for an inexperienced law student to read source documents and write the summary.
Summarization of more developed cases requires more time---around two to four hours. 
Even an experienced attorney might spend ten or more hours to understand and summarize an unusually complex case.

Figure~\ref{fig:summary-writing} illustrates the summary writing pipeline. 
After receiving a specific lawsuit assignment, the summary writer reads through court documents, especially the docket, which contains a chronological list of every document filed (Appendix~\ref{sec:guidelines-reading}).
From the massive document collection, the summary writer selects a small subset of documents (on average, eight) that provide information about major events, and attaches them to the case in CRLC.

The summary writing then takes place, guided by the instructions defined in Appendix~\ref{sec:guidelines-writing}.
To ensure the coverage of the principal information in a case, writers can resort to a checklist of facts they need to include for the case.
They typically write the long summary first, and create the short and tiny versions after, with the option to refer to the source documents as well as the longer summaries.
Cases can last a long time---sometimes several decades---so the summaries may be updated with new material several or many times as the case progresses. 
Table~\ref{table:example-summary} shows summaries
for the case {\em EEOC v. Ford Motor Company}.

After a summary writer has completed a draft of a summary, another lawyer or law student with additional experience and specialized training reviews the summary for accuracy and readability. 
When needed, the reviewer edits the summary to ensure it is factually correct and conforms with the writing style guideline (Appendix~\ref{sec:guidelines-review}).

\subsection{Dataset characterization}
\label{sec:dataset-analysis}


Table~\ref{table:overall-datasets} compares key measurements between \dataset and other SDS 
and MDS 
datasets. 
We report dataset sizes and the average number of source documents per sample (which is 1 for SDS datasets). To calculate average number of words and sentences in the source document(s) and target summaries, we use the SpaCy library~\cite{Honnibal_spaCy_Industrial-strength_Natural_2020}  \texttt{en\_core\_web\_sm} model.
Finally, we provide average extractive fragment coverage, density, and compression ratio, as defined by \citet{grusky_newsroom_2018}. \dataset is distinct in that its source text and long target summaries are much lengthier than existing datasets, with the exception of BookSum\footnote{We compare with the full book summarization task in BookSum given it has the similar source and target lengths.} which has far fewer samples and focuses on the literature domain.  Long source text poses a challenge for identifying the salient information to include in the summary.

We find that \dataset's summaries have a high fraction of terms that also appear in the source, but are still abstractive.  We follow \citet{grusky_newsroom_2018}'s approach that analyzes the coverage and density based on extractive fragments, which are shared spans of tokens that can be jointly identified in the document and summary.
\dataset has the top coverage for long summaries, of 0.94, meaning that 94\% of the words in the summary can be found in the extractive fragments from the corresponding source documents.  The generally high coverage for all \dataset granularities suggests that its summaries contain fewer unsupported entities and facts compared to the datasets with lower coverage.
At the same time, the density (or average length of the extractive fragments) ranges from 2-4 for \dataset, suggesting that most of the summary sentences are not verbatim extractions from the sources and are instead abstractive.

\section{Experiments}
\label{sec:experiments}

Our experiments on \dataset focus on two questions: (1) can models generate and synthesize information from the massive source documents in MDS tasks (\taskdtl, \taskdts, and \taskdtt); and (2) can models be configured to produce summaries of the desired lengths and details for SDS tasks (\tasklts, \taskltt, and \taskstt)? 

\subsection{Experimental Setup}

We split all cases into train (70\%, 3177 samples), Dev (10\%, 454), and Test (30\%, 908). 
All cases have long summaries, and 70\% and 36\% of the cases have short or tiny summaries, respectively.  
The corresponding source and target document lengths are reported in Table~\ref{table:overall-datasets}.
Appendix~\ref{sec:train-test-split} provides extra details about split sizes and how the splits are determined. 
\begin{table}[t]
    \caption{Performance of baseline models on different MDS tasks in \dataset.} 
    \small
    \label{table:results-mds}
    \setlength{\tabcolsep}{2.5pt}
    \renewcommand{\arraystretch}{1.25}
    \resizebox{1.\linewidth}{!}{
    \begin{threeparttable}
        \begin{tabular}{rrrrrrlrrrrrlrrrrr}
            \toprule
            \multicolumn{1}{c}{}  & \multicolumn{5}{c}{\taskdtl} & & \multicolumn{5}{c}{\taskdts} &  & \multicolumn{5}{c}{\taskdtt} \\
            \cmidrule{2-6} \cmidrule{8-12} \cmidrule{14-18}
            \textbf{Models}        & \textbf{$\text{R-1}_{f1}$} & \textbf{$\text{R-2}_{f1}$} & \textbf{$\text{R-L}_{f1}$} & \textbf{$\text{BS}_{f1}$} & \textbf{Words} &                     & \textbf{$\text{R-1}_{f1}$} & \textbf{$\text{R-2}_{f1}$} & \textbf{$\text{R-L}_{f1}$} & \textbf{$\text{BS}_{f1}$} & \textbf{Words} &                     & \textbf{$\text{R-1}_{f1}$} & \textbf{$\text{R-2}_{f1}$} & \textbf{$\text{R-L}_{f1}$} & \textbf{$\text{BS}_{f1}$} & \textbf{Words} \\
            \midrule
            \textbf{First $k$ Sent} & 30.41 & 9.67 & 14.50 & 9.73 & 813.7 &  & 21.97 & 7.17 & 13.61 & -1.60 & 120.5 &  & 8.69 & 1.07 & 6.81 & -8.72 & 33.9 \\
            \textbf{Random $k$ Sent} & 35.62 & 9.15 & 13.78 & 10.91 & 838.7 &  & 24.37 & 3.79 & 12.92 & 5.28 & 141.4 &  & 10.39 & 0.84 & 7.78 & 0.91 & 45.9 \\
            \textbf{BERT-EXT}      & 36.94 & 10.65 & 14.43 & 12.85 & 860.3 &  & 24.79 & 5.48  & 13.89 & 4.49 & 137.3  &  &  9.74 & 1.12 & 7.50  & -5.43 & 39.3 \\
            \midrule
            \textbf{PEGASUS}         & 40.79 & 20.01 & 25.36 & 34.83 & 203.8 &  & 43.35 & 19.91 & 29.99 & 37.88 & 94.6  &  & 22.61 & 7.09 & 18.44 & 26.78 & 22.3 \\
            \textbf{BART}      & 48.79 & 23.78 & 28.73 & 39.55 & 351.3 &  & 43.55 & 19.98 & 29.84 & 37.41 & 114.0 &  & 23.28 & 8.19 & 19.36 & 27.84 & 21.7 \\
            \midrule
            \textbf{LED-4096}  & 47.75 & 24.13 & 28.89 & 39.10 & 295.0 &  & 45.44 & 21.00 & 30.99 & 39.33 & 103.1 &  & 26.52 & 8.92 & 21.47 & 30.01 & 22.4   \\
            \textbf{LED-16384}    & 49.07 & 25.17 & 29.40 & 40.05 & 310.1 &  & 46.54 & 22.08 & 31.91 & 40.00 & 108.0 &  & 28.11 & 9.84 & 22.88 & 31.41 & 21.5 \\
            \textbf{PRIMERA} & 53.73 & 27.32 & 30.89 & 42.01 & 416.3 &  & 45.51 & 21.04 & 30.81 & 39.32 & 110.2 &  & 27.56 & 9.26 & 21.84 & 30.40 & 27.9 \\
            \bottomrule
        \end{tabular}    
    \end{threeparttable}
    }
    \vspace{-1mm}
\end{table}

\paragraph{Models} 
We experiment with summarization models that are representative of the state-of-the-art.
BART~\cite{lewis2019bart} and PEGASUS~\cite{zhang2020pegasus} are two recent abstractive summarizers based on the Transformer architecture~\cite{vaswani2017attention} and have achieved state-of-the-art performance on multiple summarization datasets. 
Owing to the large multi-document source content in \dataset , we also experiment with two recent summarizers tailored to this setting.
Longformer Encoder and Decoder (LED)~\cite{beltagy2020longformer} and PRIMERA~\cite{xiao2021primer} are two models that can handle longer inputs (16384 and 4096 tokens, respectively) by introducing sparsity into attention layers, and PRIMERA adds an MDS-specific pre-training objective to improve performance on MDS tasks.

\paragraph{Implementation and Computational Resources} 
For abstractive summarizers, we finetune the models based on the PyTorch~\cite{paszke2019pytorch} implementations from the HuggingFace library~\cite{wolf2019huggingface}. 
For each task, the models are trained for 6 epochs on two RTX A6000 GPUs from an internal cluster, with a learning rate of 5e-5.\footnote{For LED training, we use 3e-5 following authors' recommendations~\cite{beltagy2020longformer} and gradient checkpointing~\cite{chen2016training} to reduce GPU memory usage.} 
Following previous work~\cite{kryscinski_booksum_2021}, we use beam search with 5 beams and n-gram repetition blocks for n\textgreater 3 when decoding the generation outputs. 
The total GPU hours used for training all the benchmark models sum to roughly 300 hours.

\paragraph{Automatic Evaluation} Rouge-\{1,2,L\}~\cite{lin-2004-rouge} and BERT Score (BS)~\cite{zhang2019bertscore} are used to compute the lexical and estimated semantic overlap between the generated and gold summaries. 
We use the DeBERTA~\cite{he2020deberta} model for sentence embedding following the authors' suggestion. 
We report the average of F1 measures for Rouge and BS on the test set, and include the number of generated words for reference.

\subsection{Multi-doc legal case summarization}
\label{sec:exp-mds}

Table~\ref{table:results-mds} lists model performances on the three MDS tasks, in which the summarizers are challenged to fetch key information from the extraordinarily long input strings.  
We test a set of extractive baselines: following \citep{see_get_2017}, we develop two extractive heuristics that select the first $k$ or random $k$ sentences from the source documents ($k=35, 6, 2$ for $\textsc{l}, \textsc{s}, \textsc{t}$, respectively).
We compare them with the BERT-Extractive-Summarizer (BERT-EXT)~\cite{miller2019leveraging}, which embeds the source documents using sentence Transformers~\cite{reimers-2019-sentence-bert} and selects $k$ key sentences from the embedding clusters.
The best performing extractive models are worse than the abstractive counterparts (PEGASUS) by 47\%, 64\%, 84\% in terms of Rouge-2, and the magnitude increases as the target is more abstractive.
Because the sentence embedder is not trained for legal text, BERT-EXT attains similar (or worse in the case of \taskdts) performance to the two extractive heuristics.

For abstractive summarizers, models that allow long inputs (LED and PRIMERA) perform better than BART and PEGASUS (with only 1024 input tokens at most) on all three tasks, 
indicating the helpfulness of the longer input context. 
Because LED and PRIMERA models provide pre-trained weights with different max input lengths (16384 and 4096, respectively), we test two variants of LED (LED-16384 and LED-4096) with the corresponding input lengths.
The longer input length brings consistent performance improvements for LED across the three tasks, ranging from +4\% to +10\% of Rouge-2 in \taskdtt. 
PRIMERA outperforms even the LED-16384 model on the \taskdtl task, but achieves similar results as LED-4096 on the other two tasks of shorter targets, aligned with the authors' observation~\cite{xiao2021primer}. 

All the summarizers fail to generate long summaries of lengths that match the human summaries---PRIMERA produces the longest summaries of 416 words on average, less than 65\% of the ground-truths' average length of 647---while their generations for short and tiny summaries can match the gold label lengths (130 and 25 words on average). 
This highlights the limitations of existing summarizers in producing long abstractive summaries, as required for \dataset.

\subsection{Generating shorter summaries from the longer version}
\label{sec:exp-sds}

To further explore the multiple granularities of summary in \dataset, we train summarizers to generate shorter summaries from the longer versions.
Shown in Table~\ref{table:results-sds}, models trained on the \tasklts, \taskltt, and \taskstt task show significant improvements compared to their $\textsc{d}\rightarrow *$ counterparts: for example, the Rouge-2 of PEGASUS is improved by 79\%, 104\%, and 128\%, respectively, and exceeds scores from ``long-input'' models like LED and PRIMERA. 
The model performance in \taskstt is better than \taskltt,
providing further evidence that inputs with more condensed information simplify the summarization task.

The high summary quality when condensing long summaries to shorter ones suggests a strategy for leveraging training summaries at multiple granularities---a pipeline approach where one model generates a long summary, which is used as input in another model to generate a shorter summary. 
As an initial test, we train PRIMERA models for each of $\{\textsc{l}, \textsc{d}\}\rightarrow\textsc{s}$ and $\{\textsc{s}, \textsc{d}\}\rightarrow\textsc{t}$, which generate a short/tiny summary based on the corresponding long/short summary and the source documents.
We use ground-truth summaries $\textsc{l}$ and $\textsc{s}$ for training.
Illustrated in Table~\ref{table:results-pipeline}, when also provided with the gold long/short summaries in the input at test time, PRIMER matches the performance of the counterparts reported in \ref{table:results-sds}. 
However, when we use the model in a pipeline that does not assume a ground truth summary as input, substituting it with a BART-generated one, the performance degrades and can be worse than the corresponding $\textsc{d}\rightarrow *$ models by more than 20\% (when generating tiny summaries $\textsc{t}$). 

\begin{table}[t]
    \caption{Model performance for generating shorter summaries from the longer version.  Comparing with Table \ref{table:results-mds}, performance is much higher when the model is given a ground truth summary of a different size as input.}
    \label{table:results-longer-to-shorter}
    \begin{subtable}[t]{0.49\textwidth}
        \setlength{\tabcolsep}{4pt}
        \resizebox{1.\linewidth}{!}{
        \begin{threeparttable}
            \caption{Fine-tuning PEGASUS and BART on SDS tasks.}
            \label{table:results-sds}
            \begin{tabular}[t]{rrrrrr}
                \toprule
                \textbf{Models} & \textbf{$\text{R-1}_{f1}$} & \textbf{$\text{R-2}_{f1}$} & \textbf{$\text{R-L}_{f1}$} & \textbf{$\text{BS}_{f1}$} & \textbf{Words} \\
                \midrule
                \multicolumn{6}{c}{\tasklts Fine-tuning} \\
                \midrule
                \textbf{PEGASUS} & 54.32   & 35.62   & 42.58   & 47.49   & 156.8   \\
                \textbf{BART}    & 56.04   & 37.02   & 44.16   & 49.19   & 133.8   \\
                \midrule
                \multicolumn{6}{c}{\taskltt Fine-tuning} \\
                \midrule
                \textbf{PEGASUS} & 32.86   & 14.44   & 27.20   & 34.62   & 24.6    \\
                \textbf{BART}    & 31.65   & 13.05   & 25.52   & 33.59   & 24.0    \\
                \midrule
                \multicolumn{6}{c}{\taskstt Fine-tuning} \\
                \midrule
                \textbf{PEGASUS} & 34.15   & 16.15   & 28.27   & 34.73   & 25.6    \\
                \textbf{BART}    & 34.02   & 15.20   & 27.96   & 35.48   & 24.4    \\
                \bottomrule
                \end{tabular}
        \end{threeparttable}
        }
    \end{subtable}
    \begin{subtable}[t]{0.5\textwidth}
        \resizebox{1.\linewidth}{!}{
        \setlength{\tabcolsep}{4pt}
        \begin{threeparttable}
            \caption{PRIMERA models results on progressive summarization.}
            \label{table:results-pipeline}
            \begin{tabular}[t]{lrrrrr}
                \toprule
                \textbf{Target} & \textbf{$\text{R-1}_{f1}$} & \textbf{$\text{R-2}_{f1}$} & \textbf{$\text{R-L}_{f1}$} & \textbf{$\text{BS}_{f1}$} & \textbf{Words} \\
                \midrule
                \multicolumn{6}{c}{$\{\textsc{l}, \textsc{d}\}\rightarrow\textsc{s}$} \\[0.275mm]
                \midrule
                \textbf{Gold} $\textsc{l}$       & 54.99 & 36.42 & 43.44 & 48.69 & 133.4 \\
                \textbf{Predicted} $\textsc{l}'$ & 41.41 & 18.24 & 27.53 & 34.04 & 164.0 \\
                \midrule
                \multicolumn{6}{c}{$\{\textsc{l}, \textsc{d}\}\rightarrow\textsc{t}$} \\[0.275mm]
                \midrule
                \textbf{Gold} $\textsc{l}$       & 34.07 & 14.84 & 27.74 & 36.13 & 24.13 \\
                \textbf{Predicted} $\textsc{l}'$ & 23.63 & 7.98  & 19.50 & 27.09 & 24.05 \\
                \midrule
                \multicolumn{6}{c}{$\{\textsc{s}, \textsc{d}\}\rightarrow\textsc{t}$} \\[0.275mm]
                \midrule
                \textbf{Gold} $\textsc{l}$       & 34.60 & 16.50 & 28.71 & 35.62 & 28.65 \\
                \textbf{Predicted} $\textsc{l}'$ & 22.50 & 6.79  & 18.01 & 25.88 & 27.86 \\
                \bottomrule
                \end{tabular}
        \end{threeparttable}
        }
    \end{subtable}
    \vspace{-4mm}
\end{table}

\subsection{Multitask training for summaries of different lengths} 
\label{sec:exp-multitask}

Another strategy for leveraging summaries at multiple granularities is to train \emph{one} model that can create summaries of different lengths. 
We indicate the desired summary using prefixes~\cite{raffel2019exploring}, prepending one of ``summary: long'', ``summary: short'', or ``summary: tiny'' to the input source when for generating $\textsc{l}$, $\textsc{s}$, or $\textsc{t}$ summaries, respectively.
Table~\ref{table:exp-multitask} compares this multitask model with its single task counterparts.  We fix the same number of training epochs, thus multitask models are trained for more steps; however, increasing the number of steps was not found to improve the single-task models.

We find training for three rather than two different tasks generally leads to better performance. 
The added training samples bring greater performance boosts 
for $\textsc{t}$ summarization, which has only a third of the training samples compared to $\textsc{l}$.
$\textsc{s}$ summary results are not improved much over single-task when using three tasks, and are slightly worse using two tasks.
Most interestingly, $\textsc{l}$ summarization is greatly improved (by 11-17\% in the automated metrics) in the three-task case.
Since all training cases in the dataset have a long summary, the only difference in the multi-task training is that the model is exposed to the short and tiny views of the summary for the same cases.

\section{Human evaluation}
\label{sec:human-eval}

To assess the usability of these models, we conduct an evaluative study with law students trained to contribute summaries to the CRLC. In all, despite iterative efforts to improve system performance, we found today's models struggle to perform the task well.

\paragraph{Study Design} In coordination with the CRLC, we developed the following study setting.  (1) We scoped to only the \taskdtl setting, which is the most effort-intensive and could benefit the most from model-backed assistance.  
(2) We used a BART model to generate summaries. 
(3) Participants included two CLRC writers who edited the generations to produce summary text for 40 new cases that aren't present in \dataset; 
this process took them 180 hours in total. (4) We recorded edits made to the summaries as well as asked writers to rate the generation on a 4-point scale.\footnote{The rating levels were 0 (bad; completely unusable); 1 (somewhat helpful but requiring $>$50\% edits); 2 (requiring $<$50\% edits); 3 (perfect; no edits needed).}
\begin{table}[t]
    \caption{Comparing BART performance under multitask and single-task scenarios.  The three-task model improves performance over single-task models.}
    \label{table:exp-multitask}
    \setlength{\tabcolsep}{5pt}
    \renewcommand{\arraystretch}{1.15}
    \resizebox{1.\linewidth}{!}{
    \begin{threeparttable}
        \begin{tabular}{rrrrrrrrrrrrrr}
            \toprule
              \textbf{Lengths} &
              \textbf{Samples} &
              \textbf{$\text{R-1}_{f1}$} &
              \textbf{$\text{R-2}_{f1}$} &
              \textbf{$\text{R-L}_{f1}$} &
              \textbf{$\text{BS}_{f1}$} &
              \textbf{Words} &
               &
              \textbf{Samples} & 
              \textbf{$\text{R-1}_{f1}$} &
              \textbf{$\text{R-2}_{f1}$} &
              \textbf{$\text{R-L}_{f1}$} &
              \textbf{$\text{BS}_{f1}$} &
              \textbf{Words} \\
            \midrule
              & \multicolumn{6}{c}{\textbf{BART, Multitask:} $\textsc{d}\rightarrow\{\textsc{l}, \textsc{s}, \textsc{t}\}$} 
              &  
              & \multicolumn{6}{c}{\textbf{BART, Single-task:} $\textsc{d}\rightarrow\textsc{l}, \textsc{d}\rightarrow\textsc{s}, \textsc{d}\rightarrow\textsc{t}$} \\
            \cmidrule{2-7} \cmidrule{9-14}
            L & 6517 & 47.89      & 23.24      & 28.31     & 39.16     & 336.6     &  & 3177  & 40.79       & 20.01       & 25.36      & 34.83      & 203.8      \\
            S & 6517 & 43.80      & 20.14      & 29.89     & 38.00     & 122.6     &  & 2210  & 43.35       & 19.91       & 29.99      & 37.88      & 94.6       \\
            T & 6517 & 25.38      & 8.92       & 20.91     & 29.11     & 23.1      &  & 1130  & 22.61       & 7.09        & 18.44      & 26.78      & 22.3       \\
            \midrule  
              & \multicolumn{6}{c}{\textbf{BART, Multitask:} $\textsc{l}\rightarrow\{\textsc{s}, \textsc{t}\}$}    
              &  
              & \multicolumn{6}{c}{\textbf{BART, Single-task:} $\textsc{l}\rightarrow\textsc{s}, \textsc{l}\rightarrow\textsc{t}$}       \\
            \cmidrule{2-7} \cmidrule{9-14}
            S & 3340 & 55.20      & 36.11      & 43.42     & 48.53     & 133.5     &  & 2210   & 56.04       & 37.02       & 44.16      & 49.19      & 133.8      \\
            T & 3340 & 32.51      & 13.68      & 26.46     & 35.22     & 23.1      &  & 1130   & 31.65       & 13.05       & 25.52      & 33.59      & 24.0       \\
            \bottomrule
        \end{tabular}
    \end{threeparttable}
    }
\end{table}

\paragraph{System Design} Initial feedback from CRLC experts indicated that the end-to-end generated output of long summaries were too far from usable. Notably, they tended to hallucinate key information (e.g., filing date or court's name), and the experts stated it would take longer to correct errors than to write the summary from scratch. So we designed an alternative system based on iterative CRLC expert feedback. System features included (1) a tool for writers to select relevant text snippets while reading source documents, to aid the model in salient information selection and (2) model generation of each summary paragraph separately based on selected snippets.
Given that this system was developed in conjunction with CRLC stakeholders and greatly simplified the computer task to improve performance, we view it as a more accurate reflection of how modern summarization methods might be used in real-world applications. It thus serves as a reasonable tool to assess the usability of these models. Further details about this system can be found in the Appendix~\ref{sec:user-study-system}.


\paragraph{Results} Comparing the generations to post-edited summary texts, Rouge-1, Rouge-2, Rouge-L and BERT-Scores were 45.6, 30.0, 35.4 and 38.0, respectively; these scores are similar to those of BART from our \taskdtl experiments presented in \ref{table:results-mds}. Yet, the system generations received a 0.43 user rating, demonstrating the significant limitations of automated performance metrics. Writers averaged 87 token edits per paragraph, 76\% the average length of paragraphs, and they on average extend generation lengths by 65\%. Follow-up interviews indicated the problem of erroneous or missing key fields continued to prevent the generations from being useful.

\section{Conclusion}
\label{sec:conclusion}

In this paper, we introduce \dataset, an abstractive summarization dataset for large-scale civil rights lawsuits from U.S federal courts. 
\dataset is packed with unique features, including summaries of multiple levels of granularity for the same source, large collections of long source documents, and expert-authored summaries. 
Through a series of experiments, we find existing summarization models struggle to produce the summaries directly from the long source documents.
The average rating of 0.43 on a 0-3 scale from human assessments of current models also suggests substantial room for improvement.

\dataset is not without its limitations.
CRLC is more likely to include cases where the plaintiff wins because such cases typically last longer and receive more attention. 
This project is further limited to federal cases for which dockets are available online.
Performance might not generalize to under-represented cases (e.g., where the defendant wins); we additionally provide case metadata 
to facilitate future diagnosis of this bias.

We hope \dataset will aid development of real-world summarization systems intended to assist the activities of both specialized projects like the CRLC as well as more general sites geared toward dissemination of court documents for the general public, e.g., \url{https://www.courtlistener.com/recap/}. More broadly available and up-to-date case descriptions would be of enormous assistance to reporters, advocates, and members of the general public. The benefit would be even greater for larger ``free law'' projects that post information about hundreds of thousands, rather than thousands, of lawsuits.

\section*{Acknowledgements}

We thank the following institutions and entities who generously provide the support for the curation of the underlying Civil Rights Litigation Clearinghouse data over its 15-year history, including: University of Michigan Law School; Washington University in St. Louis School of Law; Center for Empirical Research in Law; Arnold Ventures, ``Improving Criminal Justice Reformers' Use of Litigation Information, Documents, and Insights'' (2021-2023); Vital Projects Fund, ``Revamping the Civil Rights Litigation Clearinghouse'' (2021); Proteus Fund, ``Revamping the Civil Rights Litigation Clearinghouse'' (2021); National Science Foundation SES-0718831, ``The Litigation Process in Government-Initiated Employment Discrimination Suits'' (2007). The construction of the \dataset dataset was also funded in part by NSF Convergence Accelerator Award ITE-2132318. 

We thank the hundreds of law student authors of the case summaries in \dataset, listed in \url{https://clearinghouse.net/people}. We also appreciate the advice from Adam Pah, Arman Cohan, Iz Beltagy, John Giorgi, Lucy Lu Wang, Jonathan Bragg, Dan Weld, Sida Li, and Ruochen Zhang. 

\bibliographystyle{plainnat}
\bibliography{custom}

\clearpage

\appendix

\section{\dataset release}

\subsection{Accessing \dataset}

The dataset source files are stored in JSON format, and they are uploaded to Amazon S3 and can be downloaded publicly. 
Following the HuggingFace datasets library,\footnote{\url{https://github.com/huggingface/datasets}} we develop one Python script (the \verb|multi_lexsum.py| file) that handles both the downloading of the source files and loading them into easily usable format: 
\begin{lstlisting}[language=python]
from datasets import load_dataset
multi_lexsum = load_dataset("multi_lexsum.py", name="v20220616")
# Download multi_lexsum locally and load it as a Dataset object
example = multi_lexsum["test"][0] # The first instance of the test set
example["sources"] # A list of source document text for the case
for sum_len in ["long", "short", "tiny"]:
    print(example["summary/" + sum_len]) # Summaries of three lengths
\end{lstlisting}

Currently, the \verb|multi_lexsum.py| file can be retrieved from \dataset's GitHub repository: \url{https://github.com/multilexsum/dataset}.
The authors are working on incorporating the script as part of the HuggingFace datasets library to further streamline the downloading and usage of \dataset. 
We include a similar instruction on the project website, \dataseturl, which can be regularly updated to reflect the latest changes and future updates or erratum to \dataset. 

\subsection{\dataset distribution and maintenance}

\paragraph{License} 
\dataset is distributed under the Open Data Commons Attribution License (ODC-By). 
The case summaries and metadata are licensed under the Creative Commons Attribution License (CC BY-NC), and the source documents are already in the public domain. 
Commercial users who desire a license for summaries and metadata can contact \url{info@clearinghouse.net}, which will allow free use but limit summary reposting.
The authors bear all responsibility in case of violation of rights, and confirm the dataset licenses.
The corresponding code for downloading and loading the dataset is licensed under the Apache License 2.0. 

\paragraph{Hosting and maintenance} 
The authors are committed to providing long-term support for the \dataset dataset. 
At present, \dataset files are hosted on Amazon S3 by the authors themselves. 
In the event when the authors are not able to host the data, we will migrate the data to common dataset repositories (e.g., HuggingFace Datasets) and update the documentation and code. 
The authors will closely monitor the usage of the dataset, and develop necessary updates of bug fixes when needed. 

For additional details, we refer readers to the dataset documentation (or the ``datasheet'' for datasets~\cite{gebru2021datasheets}) attached in Appendix~\ref{sec:datasheet}. 

\section{\dataset summary writing and reviewing guidelines}
\label{sec:guidelines}

\subsection{Reading source documents}
\label{sec:guidelines-reading}

Different from multi-document summarization tasks in other domains~\cite{fabbri_multi-news_2019,deyoung_ms2_2021}, \dataset summary writers are required to read several long documents and distill key facts therein. 
Strategically reading these documents saves time and effort and also improves the chances of successfully extracting the important information. 

When summary writers first begin summarizing work on a case, they can orient themselves in several ways:
\begin{itemize}[leftmargin=1em]
  \item \textit{Web searches.} Summary writers are instructed to do a quick web search when starting a case. They may find news articles or blog posts that help explain what the case is about and developments in the case. They may also find websites that are updated with documents filed in the case.
  \item \textit{Recent judicial opinions.} At the beginning of opinions, judges often summarize developments in the case up to that point.
  \item \textit{Notes from other summary writers.} When someone takes on a case picked by another person, that person usually includes notes, documents, or links to websites about the case. These can provide context or background, provide part of the narrative, or indicate some important events in the case.
\end{itemize}

Summary writers next skim their case's trial court docket to get an overview. 
A docket contains a chronological list of every document that is filed with a court in a given case, whether the filer is a party, the judge(s), or someone else. 
Each row or entry of a docket typically contains a number (its position in the list), a date, a title and short description, and---in many systems, including most federal court cases since 2003---a link to the filed document. 
The description conveys what kind of document has been filed and who filed it.

Based on the description, summary writers determine  whether a docket entry relates to something important and warrants a deeper dive into the linked document. 
While there can be up to hundreds of entries in a docket, the writers are required to whittle the long list down to typically a dozen or fewer of the documents most essential for understanding the case, which constitutes the source documents in \dataset.\footnote{Some documents might be helpful for understanding the case and writing the summary but do not need to be added to the CRLC.}
Table~\ref{table:included-docs} describes the rubric for whether to include a document based on its type.

\begin{table}[t]
    \caption{Rubrics for whether to include a source document in \dataset.}
    \label{table:included-docs}
    \resizebox{1.\linewidth}{!}{
    \begin{threeparttable}
    \renewcommand{\arraystretch}{1.1}
        \begin{tabular}{ll}
        \toprule
        \multirow{8}{*}{\textbf{Always Include}}
        & The first complaint \\
        & The last amended complaint \\
        & Settlement agreements, consent decrees, and litigated decrees \\
        & Opinions \\
        & Orders granting temporary restraining orders or preliminary injunctions \\
        & Orders granting or denying class certification \\
        & Orders awarding or denying attorneys' fees \\
        & Monitor reports \\
        \midrule
        \multirow{3}{*}{\begin{tabular}[c]{@{}l@{}}\textbf{Sometimes Include} \\ \emph{Based on Writers' Judgment}\end{tabular}}
        & Amicus briefs \\
        & Orders that settle a contentious issue \\
        & Motions/briefs if there's no order/opinion explaining their resolution \\
        \midrule
        \multirow{6}{*}{\textbf{Rarely to Never Include}}
        & Answers \\
        & Amended complaints that are not the latest one \\
        & Orders that say nothing more than what's on the docket \\
        & Motions/Briefs resolved by an opinion or order\\
        & Attorney appearances \\
        & Other orders \\
        \bottomrule
        \end{tabular}
    \end{threeparttable}
    }
\end{table}

\begin{table}[t]
    \renewcommand{\arraystretch}{1.05}
    \resizebox{1.\linewidth}{!}{
    \begin{threeparttable}
        \caption{Different Types of Source Documents in \dataset}
        \label{table:document-types}
        \begin{tabular}{lcm{0.675\linewidth}}
        \toprule
        \textbf{Document Type} & \textbf{Avg. Docs Per Case} & \textbf{Description} \\
        \midrule
        \multicolumn{3}{l}{\emph{Common Document Types}} \\ \midrule
        Complaint & 1.517 (0.88)
        & The document that starts a case and will usually be the first thing filed. Plaintiffs can also file amended complaints to add or subtract parties or claims.\\ \midrule
        Opinion/Order & 3.300 (4.78)
        & Created by judges, opinions or orders memorialize rulings in the case.\\ \midrule
        \begin{tabular}[c]{@{}l@{}}Pleading/\\ Motion/\\ Brief\end{tabular} & 1.72 (7.05)
        & This broadly covers documents filed by the parties in order to make requests or explain their arguments. \\ \midrule
        \begin{tabular}[c]{@{}l@{}}Monitor/\\ Expert/\\ Receiver Report\end{tabular} & 0.221 (1.85)
        & Reports created by non-parties to help with the litigation in various ways. A ``monitor'' is a court-appointed expert, usually superintending compliance with a court order; an ``expert'' works for one or the other side, during the litigation; a ``receiver'' is an entity appointed by the court to run defendant operations because the defendant has somehow demonstrated incapacity. \\ \midrule
        Settlement & 0.501 (0.83)
        & An agreement among parties that resolves some or all of the issues in the lawsuit. \\ \midrule
        Press Release & 0.113 (0.46)
        & What it sounds like–--a press release. \\ \midrule
        Dockets & 1.089 (0.41)
        & The docket is the court’s index to everything that has happened in a case, in that court. \\ \midrule
        \multicolumn{3}{l}{\emph{Less Common Document Types}} \\ \midrule
        Correspondence & 0.03 (0.25)
        & Letters NOT directed to the court. \newline (In some jurisdictions (particularly in New York City), parties will conduct lots of litigation through letters to the court or ``letter motions''---these are classified as motions or briefs, not as correspondence,.) \\ \midrule
        \begin{tabular}[c]{@{}l@{}}Declaration/\\ Affidavit\end{tabular} & 0.065 (0.8)
        & Documents in which someone provides information under penalty of perjury. \\ \midrule
        \begin{tabular}[c]{@{}l@{}}Discovery/\\ FOIA Material\end{tabular} & 0.018 (0.3)
        & Discovery material is evidence turned over by one party to another. The Clearinghouse usually doesn't collect it, so this document type is rare. \newline FOIA materials are documents produced in response to a Freedom of Information Act (FOIA) request. These are also uncommon in the Clearinghouse. \\ \midrule
        FOIA Request & 0.003 (0.09)
        & A request for information under the Freedom of Information Act or a state equivalent. This doesn't come up much in the Clearinghouse. \\ \midrule
        \begin{tabular}[c]{@{}l@{}}Internal \\ Memorandum\end{tabular} & 0.01 (0.12)
        & An organization's internal memo (different from litigation documents with ``memorandum'' in the title). This is a rare category. \\ \midrule
        \begin{tabular}[c]{@{}l@{}}Magistrate Report/\\ Recommendation\end{tabular} & 0.044 (0.34)
        & Decisions from magistrate judges.\\ \midrule
        \begin{tabular}[c]{@{}l@{}}Statute/\\ Ordinance/\\ Regulation\end{tabular}   & 30.018 (0.56)
        & A law or rule of government entity---federal, state, city or county, or agency. This document type includes policies created by prisons, school districts, police departments, immigration authorities, etc. \\ \midrule
        Transcripts & 0.027 (0.32) 
        & Verbatim transcripts of court proceedings or depositions. \\ \midrule
        Other & 0.050 (0.38) & \\
        \bottomrule
        \end{tabular}
    \end{threeparttable}
    }
\end{table}

\subsection{Writing summaries}
\label{sec:guidelines-writing}
The written summaries generally follow the order of events, as presented by the docket. The best summaries tell the story of the court proceedings. The student writes about the case's developments, progressing through the most important docket entries. If an entry's document is also important, the writer may also summarize the contents of the document as part of the narrative.

When writing the summary, writers also have a checklist of facts that they need to include, as illustrated in Table~\ref{table:example-summary}. 

\subsection{Reviewing summaries}
\label{sec:guidelines-review}
After a summary is written, a reviewer with additional training then checks the summaries for factual accuracy. 
Reviewers may elect to go through a docket and verify that the summary includes all the important entries, but more often they just read the summary and keep an eye out for potential gaps in the narrative, and for events that are confusing or seem implausible. 

Reviewers also ensure that the writing style conforms to the general practices as described below. 
In addition to checking spelling and grammar, they look for ways to keep the writing concise and somewhat free of legal jargon. Some examples of specific improvements:
\begin{itemize}[leftmargin=1em]
  \item \textit{Overall flow.} A summary tells a story, and so writers are encouraged to avoid too much repetition in terms of sentence structures. Following the chronology presented by the docket, beginning writers often start every paragraph with ``on x date, the party did y,'' but that can make a summary less interesting and more difficult to read. In addition, while by default summaries present events in chronological order, there are circumstances in which it makes sense for the narrative to tell pieces of the story in a different order.
  \item \textit{Level of detail.} Documents that lay out the parties' initial arguments, the court's reasoning, and the outcomes of the case can be lengthy, so students need to make sure they are including enough detail for readers to understand what happened while still summarizing the documents in a concise manner.
  \item \textit{Party descriptions.} Summary writers are instructed to describe the parties beyond just their role in a case. This tends to result in including the names of organizations or a description of individuals regarding why they would be involved in the case (e.g., ``individuals incarcerated in prison'' for a case about prison conditions).
  \item \textit{Accurate terminology.} As with any discipline, words can have a particular meaning in the legal field, and so it is important with these summaries to accurately convey the events of a case. These types of fixes include using the right verbs around motions being ``filed'' and ``granted,'' as well as including the full names of courts.
  \item \textit{Avoiding legal jargon.} These summaries are read by people other than attorneys and law students, such as policymakers and reporters. This includes avoiding the idiosyncratic capitalization sometimes found in legal writing and court documents. Part of the task of summarizing a case is to translate legal technicalities into a story more suitable for a general audience.
  \item \textit{Adding references.} If a summary discusses a judicial opinion that has a formal citation (e.g., 123 F.2d 456---meaning volume 123 of the case reporter Federal Reporter, second series, page 456), writers should include the citation, so that lawyers and researchers are able to find and cite the opinion for their own purposes. Writers may also choose to add links to news articles or blog posts that help add detail to a summary should a reader want to learn more.
  \item \textit{Grammar and spelling.} In addition to correcting syntactical mistakes, reviewers also ensure that acronyms are either avoided or spelled out the first time. In addition, our style guideline requires reviewers to ensure that the summaries are written in past tense, which beginning writers may overlook because court documents and news articles may describe some events in present tense. (Writing summaries in the past tense avoids revisions in later years to change the tense.)
\end{itemize}



\section{Usability study system design}
\label{sec:user-study-system}

\begin{figure}[t]
    \centering
    \includegraphics[width=0.95\linewidth]{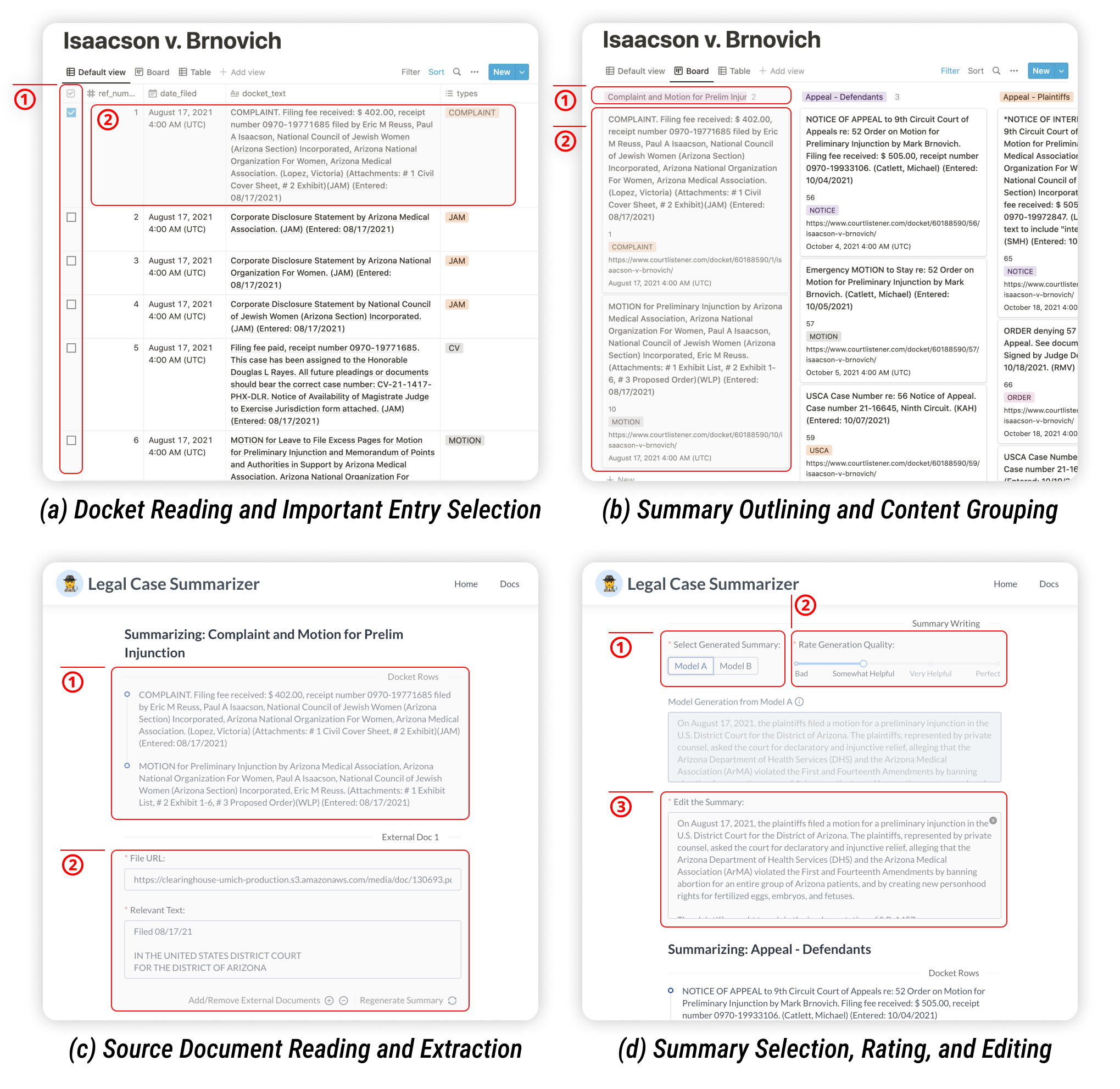}
    \caption{Illustration of the key components and functions in the user study system.} 
    \label{fig:system-illustration}

  \end{figure}

For our human evaluation of automatic summarizers applied to \dataset , we performed a usability study as described in Section \ref{sec:human-eval}. Our usability study system is designed based on iterative feedback from CRLC experts to maximize practical helpfulness. 
Illustrated in Figure~\ref{fig:system-illustration}, it breaks down case summarization into four steps: (a) docket reading and important entry selection, (b) summary outlining and content grouping, (c) source document reading and extraction, and (d) summary selection/rating/editing.\footnote{We implement the system except for the interface in (a) and (b), which is developed based on Notion: \url{https://www.notion.so/} and \url{https://github.com/lolipopshock/notion-df}.}
The system is designed to enable users to select relevant text snippets from the massive source documents to aid the model in salient information selection, and to decrease the model generation length to one paragraph at a time to reduce difficulty in both model generation as well as human editing burden. 
We detail each step:

\paragraph{(a) Docket reading and important entry selection} 
The trial court docket is the governing document in a case: it contains a chronological list of all documents filed with the court in the case, including a brief text description of each document. 
It's important for human summary authors to read through the docket and arrive at a small subset of important docket entries and documents that should be included in the summary and may warrant further reading.
This subset of candidate documents for in-depth reading are a superset of the ``source documents'' included in \dataset.
The interface in Figure~\ref{fig:system-illustration}(a) supports this step by providing a tabular interface: each row \ding{193} corresponds to an entry in the docket, and users can click the checkbox in the first column \ding{192} when the referenced document is an important event that may also need in-depth reading.

\paragraph{(b) Summary outlining and content grouping}
After reading and selecting key source documents, human authors develop an overall understanding of the case's major events, each of which will generally correspond to one paragraph in the long summary, as per the CRLC's writing guidelines.
The interface instructs writers to create ``event groups'' via the functions shown in Figure~\ref{fig:system-illustration}(b); only the candidate documents selected in the previous step are displayed in individual ``cards''.
Writers specify a group name \ding{192} and drag the relevant cards into that group \ding{193}. 
The cards can also be assigned to multiple groups when appropriate.

\paragraph{(c) Source document reading and extraction}
The writers next begin in-depth reading for documents in each event group.
In a typical unassisted workflow, authors would manually take notes consisting mainly of copied text snippets of key information from these source documents. 
Our interface, shown in Figure~\ref{fig:system-illustration}(c), includes the selected document and description from the docket in \ding{192}, and \ding{193} allows the human writers to perform this snippet-extraction workflow.

\paragraph{(d) Summary selection, rating, and editing}
Assisted by human selection of important documents and manually extracted text snippets, we run two summarizers, BART~\cite{lewis2019bart} and DistlBART~\cite{shleifer2020pre}, to produce the draft summaries. 
In Figure~\ref{fig:system-illustration}(d), the writer can pick the preferred model generation via \ding{192}, provide the 4-point rating\footnote{Rating scale described in Section~\ref{sec:human-eval}.} for the selected summary in \ding{193}, and edit the summary text in \ding{194}.

Because the models are provided human-selected salient text snippets, and are only required to generate a single paragraph at a time, the summaries generated in our setting are higher quality than those produced by end-to-end systems; this was verified via feedback from the study participants. Despite these efforts to improve generation quality, a mean rating of 0.43 for model output can be viewed as an upper-bound on the ability of modern end-to-end summarization models to produce usable summaries for this important task.



\section{Negative social impact}

We believe that release of the \dataset dataset will have positive scholarly and societal impact. However, there are some possible negatives: 

One intended use case of \dataset is to support training automatic summarizers for court documents.
However, current summarization models are known to often make up facts in the generated text~\cite{king2022don, maynez2020faithfulness}, and it is difficult to differentiate between the ``hallucinated'' and faithful information in the outputs. 
If such summarizers are deployed at scale without having solved the hallucination problem, the factually incorrect summaries could lead to misinterpretation of the case by anyone using the application. In addition, the possibility of factual errors could undermine trust in the resource even if automatic summarization is only used sparingly. 
Though this concern could be resolved by future improvements of summarization models, we highlight this risk to encourage particular care when deploying such summarizers. 

Moreover, as we discussed in Section~\ref{sec:conclusion}, the cases in \dataset are drawn from a non-representative subset of all (civil rights) cases in U.S. courts. 
Models trained on this dataset will tend to adapt the language to the cases appeared in the dataset, which could be problematic when applied to other types of cases, or to cases in different legal systems from other countries. In the short term, we acknowledge this could lead to an ``unfair'' development of NLP methods that work only for certain types of cases (though we note that the cases in question are of both particularly high public interest and, because they are non-commercial, are unlikely to spur private profit-driven development). We strongly endorse efforts to increase the transparency of the court system, including free release of court documents and case summaries for other types of lawsuits in the U.S. and for different legal systems.\footnote{For example, the SCALES-OKN project~\cite{pah2020build}.} 

\section{\dataset train-test split}
\label{sec:train-test-split}

\begin{table}[t]
    \centering
    \begin{threeparttable}
        \caption{\dataset train-dev-test splits.}
        \label{table:data-splits}
        \begin{tabular}{rrrrrr}
\toprule
                      & \textbf{Source} $\textsc{d}$ & \textbf{Long} $\textsc{l}$ & \textbf{Short} $\textsc{s}$ & \textbf{Tiny} $\textsc{t}$  & \textbf{Total} \\
\midrule
\textbf{Train (70\%)} & 28,557              & 3,177         & 2,210          & 1,130         & 6,517          \\
\textbf{Test (20\%)}  & 7,428               & 908           & 616            & 312           & 1,836          \\
\textbf{Dev (10\%)}   & 4,134               & 454           & 312            & 161           & 927            \\
\bottomrule
\end{tabular}
    \end{threeparttable}
\end{table}

\begin{table}[t]
    \centering
    \begin{threeparttable}
        \caption{The average and standard deviation of BART models' performance on different test splits.}
        \label{table:data-splits-verification}
        \begin{tabular}{cccccc}
            \toprule
            \textbf{Strategy} & \textbf{Test Split} & \textbf{$\text{R-1}_{f1}$} & \textbf{$\text{R-2}_{f1}$} & \textbf{$\text{R-L}_{f1}$} & \textbf{$\text{BS}_{f1}$} \\
            \midrule
            \multirow{3}{*}{\textbf{Hold-out}} & \textbf{10\%} & 47.21(0.34) & 23.05(0.17) & 28.21(0.34) & 38.72(0.30) \\
                                               & \textbf{15\%} & 47.00(0.13) & 22.89(0.13) & 27.94(0.14) & 38.30(0.26) \\
                                               & \textbf{20\%} & 47.25(0.02) & 23.00(0.08) & 27.99(0.10) & 38.29(0.12) \\
            \midrule
            \textbf{K-fold}                    & \textbf{20\%} & 47.24(0.49) & 23.12(0.25) & 28.23(0.29) & 38.79(0.33) \\
            \bottomrule
            \end{tabular}
    \end{threeparttable}
\end{table}

We randomly split all cases in \dataset into Train (70\%), dev (10\%) and test (20\%) sets, with detailed statistics reported in Table~\ref{table:data-splits}. 
The split strategy is verified in terms of (1) whether the test split is large enough for robust evaluation and (2) whether models are sensitive to a specific splitting of the data. 
We train and evaluate BART models under different split settings with the same hyperparameters, and we report the average and standard deviation of the ROUGE scores across different splits. 

\paragraph{Determining the optimal test split size} 
We test the same model (trained on 70\% of the data) on different test split sizes (10\%, 15\%, and 20\%), which we refer to as the ``hold-out'' strategy. 
Table~\ref{table:data-splits-verification} shows that the increased test split size leads to more stable test results (lower standard deviation).
We decide to use 20\% as the optimal test split size, while keeping 70\% of the dataset for training. 

\paragraph{Verifying model stability on different splits} 
We conduct a 5-fold cross-validation experiment, where each time the model is tested on one fold and trained on the other folds. Given our decision from the earlier step to use 70\% of the total data for training, for the purposes of this experiment, we also cap the amount of training data used in the cross validation experiments to this amount.
Shown in Table~\ref{table:data-splits-verification}, the low standard deviation indicates the models are not sensitive to specific random splits. 
Note that we used slightly different hyperparameters for this experiment than those for the main results reported in Table~\ref{table:results-mds}.

\section{\dataset datasheet}
\label{sec:datasheet}

Please see next page. 



\section*{Supplementary material}

# Multi-LexSum Dataset Sheet

We develop the dataset sheet based on the [template (v7)](#) from Gebru et al.[1] The Multi-LexSum dataset can be accessed via the *project website* https://multilexsum.github.io or the *Github Repo* https://github.com/multilexsum/dataset.

## MOTIVATION

**For what purpose was the dataset created?** Was there a specific task in mind? Was there a specific gap that needed to be filled? Please provide a description.

The Multi-LexSum dataset was curated to facilitate the development of automatic summarization methods for civil rights lawsuits.

Recent advances in document summarization have led to impressive results in generating a short description for passages typically in hundreds of words. However, the source inputs for summarizing civil right lawsuits are considerably longer: they can contain up to 70k words on average. It's still a crucial challenge for existing models to handle such long input context. Multi-LexSum is constructed to serve as a benchmark for this "long" document summarization scenario.

Additionally, human readers have different needs for summaries---ranging from one sentence to a paragraph or multi-paragraph narrations. Existing datasets only provide summaries of one granularity for a given input source, while Multi-LexSum contains summaries of up to three different levels of detailedness for one case, enabling novel research in this direction.

We also consider the summaries in Multi-LexSum to be "gold" summaries: each summary is written and reviewed by legal experts following a detailed instruction (detailed in Appendix B in the paper). In contrast, the reference summaries in previous datasets are usually obtained via automatic linking of contents, e.g., using the first sentence or summary bullets as the target summary for a piece of news, or automatically extracting and linking scientific paper abstracts and citing sentences.

**Who created the dataset (e.g., which team, research group) and on behalf of which entity (e.g., company, institution, organization)?**

The dataset is created by the collaboration between Civil Rights Litigation Clearinghouse (CRLC, from University of Michigan) and Allen Institute for AI. Multi-LexSum builds on the dataset used and posted by the Clearinghouse to inform the public about civil rights litigation.

**Who funded the creation of the dataset?** If there is an associated grant, please provide the name of the grantor and the grant name and number.

The underlying Civil Rights Litigation Clearinghouse data has been funded by numerous entities over its 15 year history, including:
- University of Michigan Law School

---

[1] Gebru, Timnit, et al. "Datasheets for datasets." Communications of the ACM 64.12 (2021): 86-92.

**Any other comments?**

None.

# COMPOSITION

**What do the instances that comprise the dataset represent (e.g., documents, photos, people, countries)?** Are there multiple types of instances (e.g., movies, users, and ratings; people and interactions between them; nodes and edges)? Please provide a description.

Each instance in the dataset represents a lawsuit and contains a set of source documents (extracted from a collection of public PDF files from U.S. federal courts), the corresponding summaries manually written by legal experts, and metadata describing attributes about the lawsuit.

**How many instances are there in total (of each type, if appropriate)?**

There are a total of 4,539 instances in this dataset, along with 40,119 source documents and 9,280 summaries. A detailed breakdown can be found in Appendix E of the paper.

**Does the dataset contain all possible instances or is it a sample (not necessarily random) of instances from a larger set?** If the dataset is a sample, then what is the larger set? Is the sample representative of the larger set (e.g., geographic coverage)? If so, please describe how this representativeness was validated/verified. If it is not representative of the larger set, please describe why not (e.g., to cover a more diverse range of instances, because instances were withheld or unavailable).

The dataset is a sample of instances, i.e., sampled from all civil rights lawsuits. It is not random; the CRLC includes cases only if they are (a) injunctive (that is, seeking court-ordered behavior/policy change) or (b) class-actions, or effectively similar to class actions (that is, adjudicating the rights of groups of people), and only in certain topics (for a list, see https://clearinghouse.net/case-types). In addition, for this project, the dataset is limited to (a) federal cases with (b) computerized dockets and documents (in a small number of cases, docket coverage may begin mid-case; these will be flagged in a planned update to the dataset). In addition, because there is no reliable way to locate every case that meets the above criteria, the sample is non-representative even among cases that fit CRLC's inclusion rules: CRLC is more likely to include cases where the plaintiff wins because such cases typically last longer and receive more attention. (As such, we clarify in the paper that performance might not generalize to under-represented cases [e.g., where the defendant wins].) We additionally provide case metadata to facilitate future diagnosis of this bias.

**What data does each instance consist of?** "Raw" data (e.g., unprocessed text or images) or features? In either case, please provide a description.

Each instance contains the following data:
1. Source documents text for a case. The text is extracted from the source PDF documents. We include the title and the type of the documents as well.
2. Summaries for a case, which can come in up to three lengths:

a. Long summaries typically contain multiple paragraphs, covering the case background, parties involved, and proceedings. Major case events and outcomes typically receive a paragraph each.
b. Short summaries have only one paragraph with a shorter description of the background, parties involved, and the outcome (so far) of the case.
c. Tiny summaries are one-sentence summaries intended to appear on Twitter to describe the case at a specific point in its history.
3. We contain the metadata for each, including but not limited to:
    a. The author(s) of the summaries
    b. Case type
    c. Case name
    d. Filing year
    e. Court & judge
    f. Plaintiff information
    g. Plaintiff attorney information
    h. Defendant information
    i. Causes of action
    j. Issue tags
    k. Prevailing party
    l. Relief information, including source and form

**Is there a label or target associated with each instance?** If so, please provide a description.

N/A.

**Is any information missing from individual instances?** If so, please provide a description, explaining why this information is missing (e.g., because it was unavailable). This does not include intentionally removed information, but might include, e.g., redacted text.

Each instance may contain summaries up to three different granularities (long, short, tiny). All the instances have a long summary, but some lack short or tiny summaries (or both). Which instances are missing short or tiny summaries is mostly a function of the date on which the summarization was done at CRLC; the summary writers were not always required to produce shorter summaries. In addition, while key metadata is available for every case, there is some missing data on less important fields, for similar reasons.

**Are relationships between individual instances made explicit (e.g., users' movie ratings, social network links)?** If so, please describe how these relationships are made explicit.

Each individual instance is considered to be independent in our dataset, though some writers may have contributed to summaries for different instances. We include an ID for the summary writer(s) for each instance.

**Are there recommended data splits (e.g., training, development/validation, testing)?** If so, please provide a description of these splits, explaining the rationale behind them.

Yes. The dataset is split into train/dev/test in the released version. We detail the statistics in Appendix E of the paper.

| **Are there any errors, sources of noise, or redundancies in the dataset?** If so, please provide a description. |
|---|
| Sometimes the source document PDFs were scanned and not digital-born. OCR was required to extract the text in these documents, which can be a source of noise. There is human error in the human summaries and human-coded metadata. |
| **Is the dataset self-contained, or does it link to or otherwise rely on external resources (e.g., websites, tweets, other datasets)?** If it links to or relies on external resources, a) are there guarantees that they will exist, and remain constant, over time; b) are there official archival versions of the complete dataset (i.e., including the external resources as they existed at the time the dataset was created); c) are there any restrictions (e.g., licenses, fees) associated with any of the external resources that might apply to a future user? Please provide descriptions of all external resources and any restrictions associated with them, as well as links or other access points, as appropriate. |
| The dataset is entirely self-contained, but it also links to CRLC, using a case-specific id. This link is not necessary (or even useful) to use the dataset. CRLC is a long-term project of the University of Michigan, funded into the future, and its pages are archived at the Internet Archive. |
| **Does the dataset contain data that might be considered confidential (e.g., data that is protected by legal privilege or by doctor–patient confidentiality, data that includes the content of individuals' non-public communications)?** If so, please provide a description. |
| There is no confidential information in our dataset; all the source documents are posted (albeit some behind a paywall) by the federal courts, available in courthouses for public inspection, and uncopyrighted and fully public. |
| **Does the dataset contain data that, if viewed directly, might be offensive, insulting, threatening, or might otherwise cause anxiety?** If so, please describe why. |
| Yes. Civil rights documents describe alleged (and eventually perhaps proven) violations of the law that harm the plaintiffs who bring the lawsuits. For example, some might concern medical neglect in prison, police violence, race or sex discrimination by employers, harmful results of abortion restrictions, and the like. In other words, the documents may contain offensive content, as they describe case allegations; these are often the central topic of the lawsuits. The data, however, should not cause any additional anxiety because every document is already made public by the federal courts, and every document and summary is already posted by CRLC. |
| **Does the dataset relate to people?** If not, you may skip the remaining questions in this section. |
| The underlying lawsuits relate to people. In addition, the summaries and metadata were written by people. |
| **Does the dataset identify any subpopulations (e.g., by age, gender)?** If so, please describe how these subpopulations are identified and provide a description of their respective distributions within the dataset. |

| |
|---|
| For cases that allege sex, race, or national origin discrimination, the affected sex, race, or national origin is coded. |
| **Is it possible to identify individuals (i.e., one or more natural persons), either directly or indirectly (i.e., in combination with other data) from the dataset?** If so, please describe how. |
| Yes. Many of the underlying cases were filed by and/or against one or more natural person; the documents name these individuals, and their lawyers. Again, all the information is already public. |
| **Does the dataset contain data that might be considered sensitive in any way (e.g., data that reveals racial or ethnic origins, sexual orientations, religious beliefs, political opinions or union memberships, or locations; financial or health data; biometric or genetic data; forms of government identification, such as social security numbers; criminal history)?** If so, please provide a description. |
| Yes, where the underlying cases relate to these issues, the documents disclose them. For example, in a case alleging race or religious discrimination, the race or religion of the plaintiff will be described in the documents (especially in documents that were filed on the plaintiff's behalf). In cases addressing criminal justice issues or jail or prison conditions, the criminal history of the plaintiffs may be relevant and discussed. However, the federal courts have rules against posting social security numbers and the CRLC has also done automated checks of the documents, as a backup, to ensure that social security numbers are not posted. |
| **Any other comments?** |
| Every case in the dataset has a docket pulled from the federal court's electronic docketing system. A very small number of cases *also* have scanned PDFs of earlier, non-digitized dockets. These are not currently tagged, but will be so in a future correction. In addition, for 118 cases, identified by metadata, the posted dockets are incomplete. |

# COLLECTION PROCESS

**How was the data associated with each instance acquired?** Was the data directly observable (e.g., raw text, movie ratings), reported by subjects (e.g., survey responses), or indirectly inferred/derived from other data (e.g., part-of-speech tags, model-based guesses for age or language)? If data was reported by subjects or indirectly inferred/derived from other data, was the data validated/verified? If so, please describe how.

For each instance in the dataset, it contains the source documents (a collection of public PDF files from U.S. federal courts) and the target summary(s). The source documents are directly observable and we developed the PDF parsers to extract the raw text from the PDF documents (detailed in PREPROCESSING/CLEANING/LABELING section). The target summaries were written and the metadata entered by legal experts summarizing the source documents following instructions.

**What mechanisms or procedures were used to collect the data (e.g., hardware apparatus or sensor, manual human curation, software program, software API)?** How were these mechanisms or procedures validated?

Manual human curation, verified by legal experts, was used. We detail the manual and summary writing guidelines in Appendix B of the paper.

**If the dataset is a sample from a larger set, what was the sampling strategy (e.g., deterministic, probabilistic with specific sampling probabilities)?**

N/A. See our discussions above in the COMPOSITION section.

**Who was involved in the data collection process (e.g., students, crowdworkers, contractors) and how were they compensated (e.g., how much were crowdworkers paid)?**

Legal experts from University of Michigan (and for earlier cases, from Washington University in St. Louis), including legal scholars, attorneys, and students. The CRLC's director is faculty and paid as such–she has not received additional compensation for her work on the Clearinghouse. CRLC has sometimes hired part-time attorneys to assist in the project; they are paid under University staff contracts. Law students have three different compensation methods: some do the work for credit; some are paid; some volunteer. For those who are paid, law students at the University of Michigan are currently paid $15/hour; the rate was a little lower in prior years. Some metadata coding of documents was performed by undergraduates; they are currently paid $12/hour, but, again, the rate was a little lower in prior years. For volunteers, the project qualifies as one of many "pro bono" projects; law students are encouraged to volunteer for such projects, and some states require several dozen hours of pro bono volunteer time as a prerequisite for attorney licensure.

| |
|---|
| **Over what timeframe was the data collected?** Does this timeframe match the creation timeframe of the data associated with the instances (e.g., recent crawl of old news articles)? If not, please describe the timeframe in which the data associated with the instances was created. |
| Multi-LexSum contains court documents from from the 1950s to 2021, heavily concentrated from 2000 to present. The contained case summaries were written between 2005 and 2021. |
| **Were any ethical review processes conducted (e.g., by an institutional review board)?** If so, please provide a description of these review processes, including the outcomes, as well as a link or other access point to any supporting documentation. |
| Schlanger explained the project to staff of the University of Michigan Institutional Review Board, and they responded that it did not need IRB approval. |
| **Does the dataset relate to people?** If not, you may skip the remaining questions in this section. |
| The dataset does not relate to people as human subjects. The cases that underlie the dataset relate to people. And people wrote the target summaries and entered the metadata. |
| **Did you collect the data from the individuals in question directly, or obtain it via third parties or other sources (e.g., websites)?** |
| All data were obtained from the CRLC, which in turn obtained documents from the federal court system and assigned legal scholars, lawyers, and law students to write the summaries. |
| **Were the individuals in question notified about the data collection?** If so, please describe (or show with screenshots or other information) how notice was provided, and provide a link or other access point to, or otherwise reproduce, the exact language of the notification itself. |
| No, individuals whose names and circumstances appear in the court documents have not been notified; the documents are public and posted by the federal courts, and are used and reposted by many sites, including CRLC. It is available to parties in court to request that documents be "sealed"--that is, made non-public. But the standards for sealing are quite stringent, because the public has a First Amendment right to know what happens in court. Similarly, parties or participants can request to proceed by pseudonym (Jane Doe, for example), but the First Amendment and policy commitments to government transparency limit such permission to cases with significant and unusual privacy interests, and to cases involving minors. In any event, the decision to file a document under seal or to seek to proceed by pseudonym is made by the affected party in court, and then adjudicated by that court. CRLC and by extension this dataset do not undertake further review.<br>The individuals who wrote the summaries and entered the metadata were notified that their summaries and work would be publicly posted (using their names) by the CRLC. This is part of appropriate acknowledgement of their work and authorship; they are not data subjects but research collaborators. Their names have long been posted at https://clearinghouse.net/people.<br>The individuals who wrote the summaries and entered the metadata were not and cannot be notified that their work are included in this dataset, because that summarization/coding was |

| |
|---|
| done over a period of more than 10 years, nearly all of it long before this dataset/development was contemplated. In any event, this dataset does not include their names, just an identifier. |
| **Did the individuals in question consent to the collection and use of their data?** If so, please describe (or show with screenshots or other information) how consent was requested and provided, and provide a link or other access point to, or otherwise reproduce, the exact language to which the individuals consented. |
| Individuals whose names and circumstances appear in the court documents have not consented separately to inclusion at CRLC or in this dataset. The documents are public and posted by the federal courts, and are used and reposted by many sites, including CRLC. However, for any who requested it over the past 15 years (by easily available email), the documents in question were either flagged using a robots.txt notice to guard against crawling, or redacted. The dataset does not include any of the flagged documents.  For any individuals identified as a summary writer by the CRLC, they agreed to participate in the project and have their authorship appropriately acknowledged and their contribution recognized during training. In any event, this dataset does not include that identification, just an id code. |
| **If consent was obtained, were the consenting individuals provided with a mechanism to revoke their consent in the future or for certain uses?** If so, please provide a description, as well as a link or other access point to the mechanism (if appropriate). |
| The authorship acknowledgement is not experimental data, but appropriate recognition of intellectual contribution. If one of the former CRLC research assistants or researchers wanted their name removed from a summary, they could reach out to CRLC and it would quickly accommodate that request. |
| **Has an analysis of the potential impact of the dataset and its use on data subjects (e.g., a data protection impact analysis) been conducted?** If so, please provide a description of this analysis, including the outcomes, as well as a link or other access point to any supporting documentation. |
| No; the authors of summaries are not data subjects but collaborators.  They are listed as such at CRLC: https://clearinghouse.net/people. |
| **Any other comments?** |
| None. |

# PREPROCESSING/CLEANING/LABELING

**Was any preprocessing/cleaning/labeling of the data done (e.g., discretization or bucketing, tokenization, part-of-speech tagging, SIFT feature extraction, removal of instances, processing of missing values)?** If so, please provide a description. If not, you may skip the remainder of the questions in this section.

Yes. We developed software to extract the text from the raw PDF files from court documents, and we store the method for extracting the document text in the released data as well.

**Was the "raw" data saved in addition to the preprocessed/cleaned/labeled data (e.g., to support unanticipated future uses)?** If so, please provide a link or other access point to the "raw" data.

Yes. The raw data will be released on a later date than the dataset release date (specified in the DISTRIBUTION section). Because the PDFs files combined are significantly larger, we are working on the best solution for long-term hosting and maintenance.

**Is the software used to preprocess/clean/label the instances available?** If so, please provide a link or other access point.

Yes. The PDF extraction code will be released in the project's Github Repo.

**Any other comments?**

None.

# USE

**Has the dataset been used for any tasks already?** If so, please provide a description.

In our paper, we demonstrate that the dataset can be used for training models to that can:
1. produce summaries for a legal case from the source documents,
2. perform controlled automatic summarization that can produce summaries of different granularities.
3. condense a long summary to a short version.

**Is there a repository that links to any or all papers or systems that use the dataset?** If so, please provide a link or other access point.

N/A. If papers are produced, links will be posted at https://clearinghouse.net/search/resources/?resource_types=6132

**What (other) tasks could the dataset be used for?**

We envision the dataset can also be used for the following scenarios, including but not limited to:
1. Large-scale pre-training for legal document understanding models. As we provide a massive collection of documents, they can be used as the pre-training corpus for large language models for understanding legal text.
2. Information extraction models from legal documents. As we provide metadata for each case (e.g., causes of actions, outcomes), and they are based on the source documents, a potential use case might be training an information retrieval model for these fields from the source documents.

**Is there anything about the composition of the dataset or the way it was collected and preprocessed/cleaned/labeled that might impact future uses?** For example, is there anything that a future user might need to know to avoid uses that could result in unfair treatment of individuals or groups (e.g., stereotyping, quality of service issues) or other undesirable harms (e.g., financial harms, legal risks) If so, please provide a description. Is there anything a future user could do to mitigate these undesirable harms?

As discussed above and in the paper, the cases in the dataset are a non-representative sample of all civil right lawsuits, with various inclusion criteria (electronic availability, topic area, injunctive and class litigation) and also some practically-produced selection bias: CRLC is more likely to include cases where the plaintiff wins because such cases typically last longer and receive more attention. As such, we clarify in the paper that performance might not generalize to under-represented cases (e.g., where the defendant wins). The dataset should not be used for training models to predict the outcome of a lawsuit.

**Are there tasks for which the dataset should not be used?** If so, please provide a description.

Given the constraints mentioned above, the dataset should not be used for training models to predict the outcome of a lawsuit.

**Any other comments?**

None.

# DISTRIBUTION

**Will the dataset be distributed to third parties outside of the entity (e.g., company, institution, organization) on behalf of which the dataset was created?** If so, please provide a description.

Yes, the dataset will be publicly available on the internet.

**How will the dataset be distributed (e.g., tarball on website, API, GitHub)?** Does the dataset have a digital object identifier (DOI)?

The dataset will be uploaded to an Amazon S3 bucket on AWS, and people can download it publicly via the provided link. In addition, we release a (Python) script for loading and using the dataset files in https://github.com/multilexsum/dataset. We plan to incorporate the dataset to the Huggingface Datasets library for easy access in the future.

**When will the dataset be distributed?**

Since June 16, 2022.

**Will the dataset be distributed under a copyright or other intellectual property (IP) license, and/or under applicable terms of use (ToU)?** If so, please describe this license and/or ToU, and provide a link or other access point to, or otherwise reproduce, any relevant licensing terms or ToU, as well as any fees associated with these restrictions.

The Multi-LexSum dataset is distributed under the Open Data Commons Attribution License (ODC-By). The case summaries and metadata are licensed under the Creative Commons Attribution License (CC BY-NC), and the source documents are already in the public domain. Commercial users who desire a license for summaries and metadata can contact info@clearinghouse.net, which will allow free use but limit summary reposting.

**Have any third parties imposed IP-based or other restrictions on the data associated with the instances?** If so, please describe these restrictions, and provide a link or other access point to, or otherwise reproduce, any relevant licensing terms, as well as any fees associated with these restrictions.

No.

**Do any export controls or other regulatory restrictions apply to the dataset or to individual instances?** If so, please describe these restrictions, and provide a link or other access point to, or otherwise reproduce, any supporting documentation.

N/A.

**Any other comments?**

None.

# MAINTENANCE

**Who will be supporting/hosting/maintaining the dataset?**

Zejiang Shen is supporting the dataset.

**How can the owner/curator/manager of the dataset be contacted (e.g., email address)?**

Zejiang Shen will be the main contact for the dataset, and the up-to-date contact information can be retrieved at [www.szj.io](www.szj.io). Questions pertaining to the Allen Institute for AI's involvement in curating or maintaining this dataset should be directed to [dougd@allenai.org](dougd@allenai.org). Questions pertaining to the CRLC's involvement in curating or maintaining this dataset and/or the CRLC's past and ongoing efforts to produce and disseminate these summaries should be directed to [info@clearinghouse.net](info@clearinghouse.net).

**Is there an erratum?** If so, please provide a link or other access point.

Currently we haven't found any errors in the version (to be released). If we do, we will post the erratum and update information in the dataset website and Github repo.

**Will the dataset be updated (e.g., to correct labeling errors, add new instances, delete instances)?** If so, please describe how often, by whom, and how updates will be communicated to users (e.g., mailing list, GitHub)?

Though we have no concrete plans for dataset updates, we envision there will be updated versions for error corrections and inclusion of additional data. If there are any updates, the updated information will be posted on the dataset website and Github repo.

**If the dataset relates to people, are there applicable limits on the retention of the data associated with the instances (e.g., were individuals in question told that their data would be retained for a fixed period of time and then deleted)?** If so, please describe these limits and explain how they will be enforced.

N/A.

**Will older versions of the dataset continue to be supported/hosted/maintained?** If so, please describe how. If not, please describe how its obsolescence will be communicated to users.

Yes. The public links of the dataset files contain version information. We will keep the links for older versions available after the release of newer versions.

**If others want to extend/augment/build on/contribute to the dataset, is there a mechanism for them to do so?** If so, please provide a description. Will these contributions be validated/verified? If so, please describe how. If not, why not? Is there a process for communicating/distributing these contributions to other users? If so, please provide a description.

Given that we released the dataset under the Creative Commons (CC BY-NC) license, others should feel free to extend and build upon the dataset. Any contributions to the dataset can happen in the form of Github Pull Requests / Issues: the contributors can submit the changes or suggestions, and we will monitor and moderate them monthly.

**Any other comments?**

None.



\end{document}